\documentclass{article}
\PassOptionsToPackage{table}{xcolor}
\usepackage{iclr2027_conference,times}
\iclrfinalcopy

\usepackage{amsmath,amsfonts,bm}

\def\eqref#1{equation~\ref{#1}}

\def\1{\bm{1}}

\DeclareMathAlphabet{\mathsfit}{\encodingdefault}{\sfdefault}{m}{sl}
\SetMathAlphabet{\mathsfit}{bold}{\encodingdefault}{\sfdefault}{bx}{n}

\DeclareMathOperator*{\argmin}{arg\,min}

\DeclareRobustCommand{\paperEqref}[1]{(\ref{#1})}
\let\eqref\paperEqref

\usepackage[T1]{fontenc}
\usepackage[utf8]{inputenc}
\usepackage{microtype}
\usepackage{graphicx}
\usepackage{amsmath,amssymb}
\usepackage{booktabs}
\usepackage{enumitem}
\usepackage[table]{xcolor}
\usepackage{colortbl}
\usepackage{adjustbox}
\usepackage{algorithm}
\usepackage{algorithmic}
\usepackage[most]{tcolorbox}
\usepackage{url}
\usepackage{hyperref}
\usepackage{multirow}
\usepackage{wrapfig}
\usepackage{needspace}
\makeatletter
\newcommand{\inlinefigurecaption}[1]{\def\@captype{figure}\caption{#1}}
\makeatother

\graphicspath{{figures/}}

\definecolor{darkblue}{RGB}{0,0,153}
\hypersetup{
  colorlinks=true,
  linkcolor=darkblue,
  citecolor=darkblue,
  urlcolor=darkblue
}

\newcommand{\papertitle}{FisherEdit: Fisher Information Guided Knowledge Editing for Large Language Models}
\title{\papertitle}
\newcommand{\authornames}{Yizhou Yang, Jiahong Fu, Zhiwu Wang, Kailong Fan, Haokun Lin, Yichen Wu, Zhenan Sun, Renzhen Wang, Deyu Meng}
\author{%
  \begin{minipage}[t]{\dimexpr\textwidth-2\tabcolsep\relax}
  \normalfont
  \centering
  \textbf{Yizhou Yang\textsuperscript{1,*}\quad
  Jiahong Fu\textsuperscript{1,*}\quad
  Zhiwu Wang\textsuperscript{1}}\\[2pt]
  \textbf{Kailong Fan\textsuperscript{2}\quad
  Haokun Lin\textsuperscript{3,$\dagger$}\quad
  Yichen Wu\textsuperscript{4,$\dagger$}}\\[2pt]
  \textbf{Zhenan Sun\textsuperscript{5}\quad
  Renzhen Wang\textsuperscript{1,$\dagger$}\quad
  Deyu Meng\textsuperscript{1}}\\[6pt]
  {\small
  \textsuperscript{1}Xi'an Jiaotong University\\
  \textsuperscript{2}Zhejiang University\\
  \textsuperscript{3}City University of Hong Kong\\
  \textsuperscript{4}Harvard University\\
  \textsuperscript{5}Institute of Automation, Chinese Academy of Sciences\\[4pt]
  \textsuperscript{*}Equal contribution.\quad
  \textsuperscript{$\dagger$}Corresponding author.}
  \end{minipage}%
}
\date{}

\hypersetup{
  pdftitle={\papertitle},
  pdfauthor={\authornames},
  pdfsubject={Knowledge editing for large language models}
}

\begin{document}
\maketitle
\fancyhead{}
\renewcommand{\headrulewidth}{0pt}
\begin{abstract}
Sequential knowledge editing in Large Language Models (LLMs) often leads to a degradation of general capabilities due to cumulative parameter interference. In this paper, we revisit prominent projection-based methods like AlphaEdit and reveal a surprising mismatch: their robustness is primarily driven by update regularization rather than the intended null-space projection. Based on this insight, we propose FisherEdit, a theoretically grounded method built on Fisher geometry, which replaces coarse activation-space projections with importance-aware regularization in parameter space. It penalizes changes to sensitive parameters more strongly while retaining flexibility for new facts, and admits an efficient row-wise closed-form solution. Experiments on Llama-3, Qwen2.5, and Olmo-3 across multiple benchmarks demonstrate that FisherEdit achieves state-of-the-art performance while better preserving model stability and downstream task accuracy.
\end{abstract}

\begin{figure}[h]
    \centering
    \includegraphics[width=0.85\linewidth]{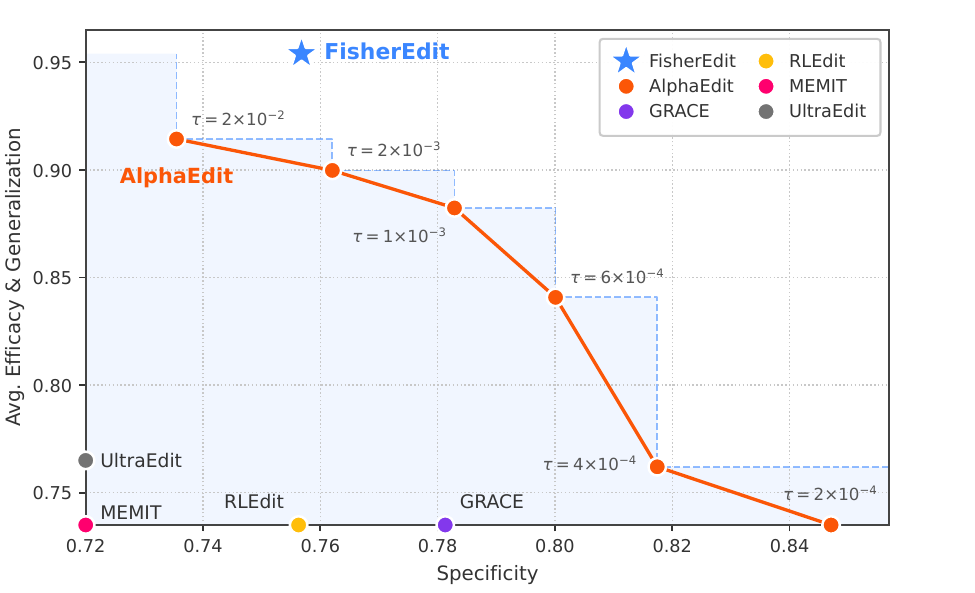}
    \caption{Efficacy--locality trade-off on Llama-3-8B-Instruct. Sweeping AlphaEdit's truncation threshold $\tau$ traces a monotone Pareto curve. FisherEdit lands strictly above the curve.}
    \label{fig:pareto}
\end{figure}

\section{Introduction}
\label{sec:intro}

Large language models (LLMs) encode extensive factual knowledge during pretraining, and this knowledge must often be revised as facts become outdated, incorrect, or inconsistent with new evidence~\citep{petroni2019language,roberts2020much,geva2021transformer,xia2025medrek,shi2025dualedit}. Knowledge editing addresses this need by updating specific facts without full retraining, while aiming to preserve unrelated behavior~\citep{yao2023editing,wang2024knowledge}. {This goal is especially difficult in sequential editing, where repeated weight changes can accumulate, interfere with one another, and degrade the model's general abilities}~\citep{yang2024butterfly,wu2024mitigating}. Maintaining reliable behavior under repeated edits, therefore, remains a central challenge for knowledge editing.

A prominent approach is direct weight editing. Locate-then-edit methods~\citep{meng2022locating,meng2022mass} {identify where factual associations are stored and update the corresponding weights.} Because repeated updates can interfere with previously edited or preserved knowledge, recent methods add preservation constraints to restrict how the model changes. AlphaEdit~\citep{fang2025alphaedit} is a representative example: it uses a null-space projection to steer edits away from activation directions associated with prior knowledge. This projection-based view has become influential in sequential editing and has motivated several follow-up methods~{\citep{xu2025knowledge,ikram2026crispedit}}.

\begin{figure}[t]
    \centering
    \includegraphics[width=\linewidth]{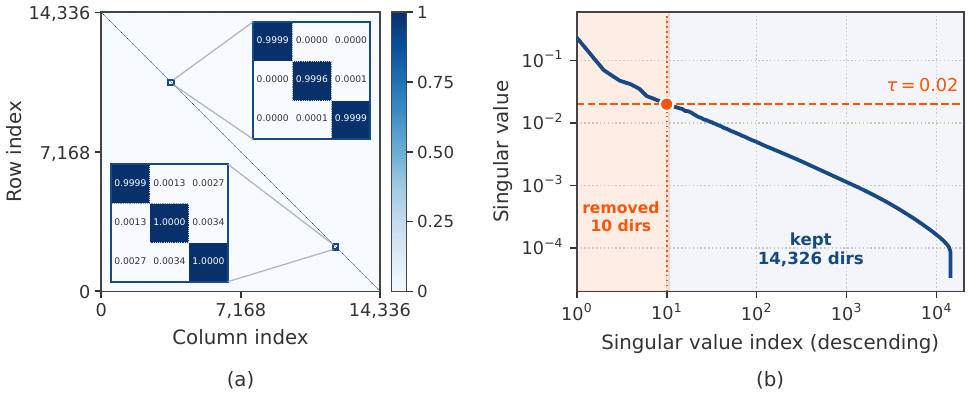}
    \caption{(a) AlphaEdit's null-space projector {\small$\mathbf{P}_\tau$} on Llama-3-8B-Instruct is close to the identity matrix, indicating that most edit directions are only weakly affected by the projection. (b) Singular spectrum of {\small $\mathbf{K}_0 \mathbf{K}_0^\top$} on the same layer. Only a few dozen directions exceed the truncation threshold $\tau$.}
    \label{fig:p_identity_heatmap}
\end{figure}

{Given AlphaEdit's influence, we revisit it to examine what its null-space projection actually contributes.} AlphaEdit offers a clean geometric intuition: edits are projected away from activation directions associated with prior knowledge, so the corresponding knowledge should be preserved.
{However, this intuition only partially explains its empirical behavior.}
As illustrated in Fig.~\ref{fig:p_identity_heatmap}, the default projector remains close to the identity and removes only a small portion of the activation space, leaving most update directions largely unchanged. Our more systematic analysis further shows that preservation in AlphaEdit is strongly shaped by update regularization, not projection alone. This motivates us to rethink preservation in sequential editing as a problem of designing more precise update regularization.

The key issue is how to constrain edits without removing the flexibility needed to encode new facts. Existing preservation mechanisms often operate at a coarse level: hard subspace constraints remove entire directions~{\citep{lyu2025evoedit,xu2025knowledge,ikram2026crispedit}}, while regularization-based methods typically constrain the update through global magnitude, complexity, or spectral properties~\citep{gupta2025lifelong,liu2025energy,zhang2026spectral}. {While effective, these constraints do not directly distinguish parameters by their importance to general behavior. Parameter sensitivity in LLMs is highly uneven: some weight perturbations strongly affect general predictions, while others have limited impact}~\citep{frantar2023sparsegpt,lin2024awq}. Effective preservation should therefore be importance-aware, imposing stronger constraints on sensitive parameters while leaving less critical parameters available for editing.
{Following this principle, we propose \textbf{FisherEdit}, a Fisher-guided regularization method for sequential knowledge editing. It uses a diagonal Fisher estimate to measure parameter sensitivity and weight the update penalties accordingly. This constrains sensitive weights more strongly while allowing less sensitive weights to adapt to new facts. Our contributions are as follows:}
\begin{itemize}[leftmargin=3mm, itemsep=-1mm, topsep=-0.5 mm]
    \item {We re-examine AlphaEdit and find that its preservation benefit comes mainly from update regularization, with little contribution from the default null-space projection. Strengthening the projection trades edit efficacy for locality.}
    \item Based on our analysis, we rethink preservation in sequential
    editing as regularization and introduce FisherEdit, a theoretically
    motivated method that uses Fisher geometry to derive an importance-aware,
    element-wise update penalty, {together with a specialized row-wise
    closed-form solver.}
    \item Across backbones and benchmarks, FisherEdit achieves state-of-the-art performance while better preserving general abilities on downstream tasks and reducing representation- and parameter-space drift relative to prior methods.
\end{itemize}

\begin{figure}[!t]
    \centering
    \includegraphics[width=\linewidth]{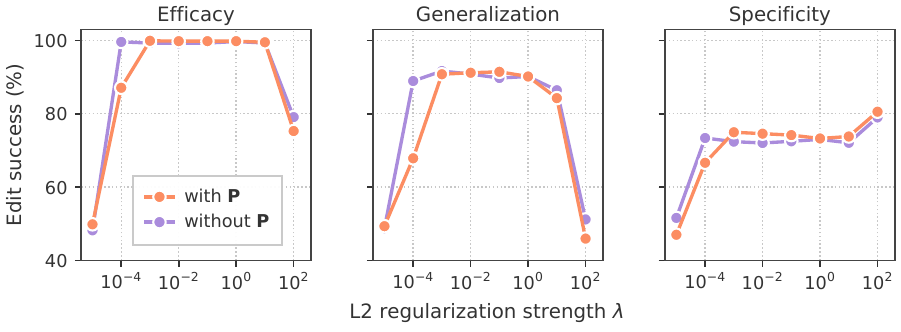}
    \caption{Edit success for AlphaEdit and projector-disabled variant on Llama-3-8B. Across most of the regularization range, the w/ and w/o {\small $\mathbf{P}$} curves almost overlap on all three edit-quality metrics.}
    \label{fig:l2_efficacy}
\end{figure}

\section{Related Work}
\label{sec:related}

{Knowledge editing (KE) updates factual associations in LLMs or reroutes queries to corrected knowledge}, without full retraining or broad degradation of unrelated behavior~\citep{yao2023editing,wang2024knowledge}. Existing methods can be broadly grouped into learning-based editors and direct weight-editing methods.

\noindent \textbf{Learning-based editors} train auxiliary modules to produce, route, or store edits for a frozen base model. Representative methods include KnowledgeEditor~\citep{de2021editing}, MEND~\citep{mitchellfast}, MALMEN~\citep{tan2024massive}, SERAC~\citep{mitchell2022memory} and RLEdit~\citep{li2025reinforced}. {These methods require additional training, and their effectiveness depends on the editor's capacity and training distribution.}

\noindent {\textbf{Direct weight-editing methods} modify target-model weights directly. Knowledge Neurons~\citep{dai2022knowledge} and ROME~\citep{meng2022locating} localize factual associations in transformer MLPs, while MEMIT~\citep{meng2022mass} {extends this to batched least-squares editing}. Subsequent work adds constraints to preserve prior knowledge~\citep{gu2024model,ma2025perturbation,zhang2025uncovering}. Among projection-based methods, AlphaEdit~\citep{fang2025alphaedit} uses the null space of preserved activation keys; EvoEdit~\citep{lyu2025evoedit} evolves it during editing; and KDE~\citep{xu2025knowledge} decouples past and current knowledge through orthogonal projection. CrispEdit~\citep{ikram2026crispedit} instead hard-thresholds a K-FAC curvature spectrum, restricting edits to the approximate null space of capability curvature. Such hard projections can remove both prior-knowledge and edit-relevant signals, trading efficacy for locality. {FisherEdit instead uses a soft, element-wise Fisher penalty, retaining all update directions while constraining important parameters more strongly.}}

\noindent {\textbf{Connection to continual-learning regularization.}
{Importance-weighted regularization is classical in continual learning:}
EWC~\citep{kirkpatrick2017overcoming} penalizes movement of
parameters with large diagonal Fisher, and SI~\citep{zenke2017continual},
MAS~\citep{aljundi2018memory}, and Meta-CL~\citep{wu2024meta} derive related online importance estimates.
{FisherEdit adapts this principle to least-squares editing, where
updates are computed in closed form rather than through SGD. Its Fisher
penalty is motivated by a local second-order approximation to predictive
KL divergence (Sec.~\ref{sec:theory}).}}

\section{Rethinking Null-Space Projection}
\label{sec:rethinking}
{We now examine what drives preservation in AlphaEdit~\citep{fang2025alphaedit}. After introducing its editing objective, we use controlled ablations to separate the effects of null-space projection and update regularization.}

\subsection{Preliminaries}
\label{sec:preli}
Following~\citep{meng2022locating,meng2022mass}, we edit the MLP down-projection matrix {\small $\mathbf{W} \in \mathbb{R}^{d \times m}$}. For a batch of $n$ edits, let {\small $\mathbf{K}\in\mathbb{R}^{m\times n}$} denote the edit keys, {\small $\mathbf{V}\in\mathbb{R}^{d\times n}$} the target values. The goal is to find an update {\small $\Delta\in\mathbb{R}^{d\times m}$} such that the edited weights map edit keys to the target values:
\begin{equation}
(\mathbf{W}+\Delta) \mathbf{K} \approx \mathbf{V} .
\end{equation}
Equivalently, $\Delta$ should fit the residual left by the current weights:
\begin{equation}
\Delta \mathbf{K} \approx \mathbf{E},
\qquad
\mathbf{E} \triangleq \mathbf{V}-\mathbf{W} \mathbf{K} .
\end{equation}

In sequential editing, the update should also avoid changing previous edits and unrelated knowledge. Let {\small $\mathbf{K}_p\in\mathbb{R}^{m\times n_p}$} denote historical edit keys from previous batches, {\small $\mathbf{K}_1\in\mathbb{R}^{m\times n_1}$} denote current edit keys, and {\small $\mathbf{K}_0\in\mathbb{R}^{m\times n_0}$} denote preserved keys sampled from a background corpus. A standard locate-then-edit objective is
{\small
\begin{equation}
\Delta
=
\argmin_{\tilde{\Delta}}
\underbrace{\|\tilde{\Delta}\mathbf{K}_1-\mathbf{E}\|^2}_{(a)}
+
\underbrace{\|\tilde{\Delta}\mathbf{K}_p\|^2}_{(b)}
+
\underbrace{\|\tilde{\Delta}\mathbf{K}_0\|^2}_{(c)} .
\label{eq:memit}
\end{equation}
}
Term (a) fits the new edits, while terms (b) and (c) penalize changes along historical-edit and preserved-knowledge directions, respectively.

\noindent \textbf{AlphaEdit}~\citep{fang2025alphaedit} replaces the penalty on {\small $\mathbf{K}_0$} with a null-space projector {\small$\mathbf{P}_\tau$}. Specifically, {\small$\mathbf{P}_\tau$} is computed from the eigenspectrum of {\small $\mathbf{K}_0\mathbf{K}_0^\top$}: directions whose eigenvalues exceed a threshold $\tau$ are treated as occupied, and the remaining subspace is used for editing. AlphaEdit solves
\begin{equation}
\Delta =
\argmin_{\tilde\Delta}
\|\tilde\Delta \mathbf{P}_\tau \mathbf{K}_1 \! - \mathbf{E}\|^2
\! + \!
\|\tilde\Delta \mathbf{P}_\tau \mathbf{K}_p\|^2
\! + \!
\lambda \|\tilde\Delta \mathbf{P}_\tau\|^2 .
\label{eq:alphaedit}
\vspace{-1mm}
\end{equation}
The following analysis asks whether the projection {\small$\mathbf{P}_\tau$}, the regularization strength $\lambda$, or the threshold $\tau$ is responsible for AlphaEdit's empirical preservation behavior.

\subsection{What Actually Drives Preservation? }
\label{sec:rethinking:f1}

\noindent \textbf{Finding 1: The Projector {\small $\mathbf{P}_\tau$} is near-identity.}
If the null-space mechanism is the main source of preservation, the projector should remove a substantial set of directions associated with prior knowledge. We first inspect the projector used by AlphaEdit under its default setting. As shown in Fig.~\ref{fig:p_identity_heatmap}(a), the matrix is almost identical to the identity: its diagonal entries are close to one, and its off-diagonal entries are negligible.
This behavior comes from the spectrum of {\small $\mathbf{K}_0\mathbf{K}_0^\top$}. Fig.~\ref{fig:p_identity_heatmap}(b) shows that only a small number of singular directions exceed the default threshold $\tau$, while the vast majority fall below it and are kept for editing. Thus, the projected-out subspace covers only a tiny fraction of the activation space. In practice, {\small $\mathbf{P}_\tau$} leaves most update directions unchanged.

We then ask whether this near-identity projector affects editing outcomes. We compare AlphaEdit with a variant that disables the projector while keeping the same regularization strength. As shown in Fig.~\ref{fig:l2_efficacy}, the two variants largely overlap across Efficacy, Generalization, and Specificity metrics over the working range of $\lambda$. This suggests that, under the default setting, AlphaEdit's projector contributes little to its empirical preservation behavior.

\noindent \textbf{Finding 2: Regularization is the main control knob.}
If the projector changes little, the remaining candidate explanation is the update regularizer. To isolate its effect, we keep the projector fixed and vary only the regularization strength $\lambda$. Fig.~\ref{fig:l2_lambda_bars} shows that this single change produces large shifts across all edit-success metrics. With weak regularization, the update becomes unstable and performance collapses; with stronger regularization, performance is substantially restored.
\noindent{These results suggest that much of AlphaEdit's robustness comes from controlling update magnitude through regularization, with limited contribution from the default projection.}

\par\addvspace{6pt}
\noindent\begin{minipage}[t]{0.49\linewidth}
\vspace{0pt}
\noindent\textbf{Finding 3: Stronger projection only moves along a trade-off.} One possible explanation is that the default threshold $\tau$ makes the projector too weak. We therefore lower $\tau$, forcing {\small $\mathbf{P}_\tau$} to remove more activation directions. If the null-space interpretation were sufficient, a stronger projection should improve preservation without harming edit success. The results show a different pattern. As $\tau$ decreases, specificity improves, but efficacy and generalization drop sharply (Fig.~\ref{fig:pareto}). Thus, stronger projection does not recover a better protected subspace; it simply trades editing success for locality. This happens because the removed directions are not purely preserved-knowledge directions. They can also contain information needed to realize the edit. Removing them therefore protects some unrelated behavior, but weakens the edit itself.
\end{minipage}\hfill
\begin{minipage}[t]{0.46\linewidth}
    \vspace{0pt}
    \centering
    \setlength{\abovecaptionskip}{3pt}
    \includegraphics[width=\linewidth]{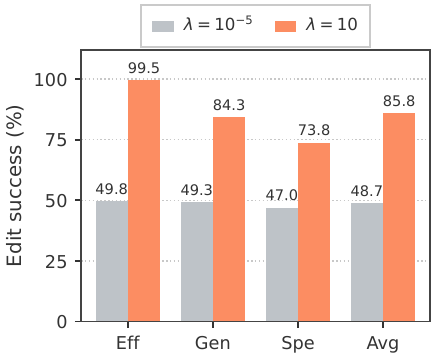}
    \inlinefigurecaption{Within AlphaEdit on Llama-3-8B-Instruct, changing $\lambda$ from $10^{-5}$ to $10$ produces a large jump in all three metrics.}
    \label{fig:l2_lambda_bars}
\end{minipage}
\par\addvspace{6pt}

\noindent \textbf{Takeaway.}
These findings suggest a different design principle for sequential editing: preservation should act on parameters, not entire activation directions. Hard projection can remove directions needed for the new edit, while uniform shrinkage ignores which weights matter most for general behavior. A better constraint should distinguish sensitive weights from safer ones. This motivates FisherEdit, which applies a Fisher-weighted, element-wise penalty directly to the update.
\par

\section{Method}
\label{sec:method}
\subsection{Fisher-Guided Editing Objective}
Motivated by the analysis above, our FisherEdit shifts preservation from activation-space projection to parameter-space regularization. We keep the standard edit-fitting objective, but penalize each parameter update according to its Fisher importance:
{\small
\begin{equation}
\Delta
=
\argmin_{\tilde{\Delta}}
\|\tilde{\Delta}\mathbf{K}_1- \mathbf{E}\|^2
+
\|\tilde{\Delta}\mathbf{K}_p\|^2
+
\lambda\|\sqrt{\mathbf{F}}\odot\tilde{\Delta}\|^2 ,
\label{eq:goal}
\end{equation}
}
Here, $\mathbf{F}\in\mathbb{R}^{d\times m}$ is the diagonal Fisher Information matrix and $\odot$ denotes element-wise multiplication. This term makes changes to high-Fisher parameters more costly while leaving low-Fisher parameters more flexible for encoding new facts.
Compared with the objectives in Eqns.~(\ref{eq:memit})(\ref{eq:alphaedit}), FisherEdit keeps the edit-fitting and historical-edit terms, but replaces the preserved-key penalty with a Fisher-weighted penalty on the update itself.

\subsection{\texorpdfstring{{Solving the Fisher-Weighted Objective}}{Solving the Fisher-Weighted Objective}}
{Eqn.~\eqref{eq:goal} is a quadratic objective that admits a closed-form solution. Because the Fisher penalty is element-wise, we can solve for each row of the update independently. Let $\mathbf{G}=\mathbf{K}_1\mathbf{K}_1^\top+\mathbf{K}_p\mathbf{K}_p^\top$, and let $\delta_i$, $e_i$, and $f_i$ denote the $i$-th rows of $\Delta$, $\mathbf{E}$, and $\mathbf{F}$, respectively. With $\mathbf{D}_i=\mathrm{diag}(f_i)$, setting the gradient of Eqn.~\eqref{eq:goal} to zero gives $\delta_i(\mathbf{G}+\lambda\mathbf{D}_i)=e_i\mathbf{K}_1^\top$ for each row, yielding
\begin{equation}
\delta_i
=
e_i\mathbf{K}_1^\top(\mathbf{G}+\lambda \mathbf{D}_i)^{-1}.
\label{eq:row_solution}
\end{equation}
The shared matrix $\mathbf{G}$ captures the constraints from current and historical edit keys, while the row-specific $\mathbf{D}_i$ weights the update according to parameter importance.}

{Directly evaluating Eqn.~\eqref{eq:row_solution} requires an $m\times m$ matrix inverse for each row. Since $m$ is the MLP intermediate dimension, this becomes expensive at LLM scale. We therefore exploit the low-rank structure $\mathbf{G}=\mathbf{U}\mathbf{U}^\top$, where $\mathbf{U}=[\mathbf{K}_1\mid\mathbf{K}_p]\in\mathbb{R}^{m\times r}$ and $r=n_1+n_p$. Applying the Woodbury identity~\citep{golub2013matrix} gives the equivalent update
\begin{equation}
\boxed{
\begin{aligned}
\delta_i
=
e_i\mathbf{K}_1^\top
\Big[
&(\lambda \mathbf{D}_i)^{-1}
-
(\lambda \mathbf{D}_i)^{-1}\mathbf{U} \\
&\left(
\mathbf{I}_r
+
\mathbf{U}^\top(\lambda \mathbf{D}_i)^{-1}\mathbf{U}
\right)^{-1}
\mathbf{U}^\top(\lambda \mathbf{D}_i)^{-1}
\Big].
\end{aligned}
}
\label{eq:efficient_row_solution}
\end{equation}
Since $\mathbf{D}_i$ is diagonal, $(\lambda\mathbf{D}_i)^{-1}$ is computed by element-wise division. The remaining linear system has size $r\times r$, reducing the cost when the number of edit keys is much smaller than the MLP dimension ($r\ll m$). All $d$ row-wise systems can be solved using batched linear algebra, with an overall complexity of $O(dmr^2+dr^3)$. The full derivation is provided in Appendix~\ref{app:row_solution_derivation}.}

\subsection{Estimating the Fisher Diagonal}
\label{sec:fisher_computation}
The solver above assumes to the Fisher diagonal $\mathbf{F}$ for the edited weight block. We estimate this quantity once before editing and reuse it throughout the edit sequence. This keeps the editing procedure efficient while giving each parameter a fixed importance score measured from the model's original predictive behavior.

We use the empirical Fisher diagonal. Given a held-out calibration set $\mathcal{D}=\{(x_t,y_t)\}_{t=1}^{N}$, the Fisher value for parameter $W^{(\ell)}_{ij}$ in layer $\ell$ is estimated as
\begin{equation}
\mathbf{F}^{(\ell)}_{ij}
=
\frac{1}{N}
\sum_{t=1}^{N}
\left(
\frac{\partial \mathcal{L}(x_t,y_t)}
{\partial W^{(\ell)}_{ij}}
\right)^2 ,
\label{eq:fisher_diag}
\end{equation}
where $\mathcal{L}$ is the cross-entropy loss. We compute this estimate only for the MLP down-projection matrices that are later edited, rather than for all model parameters.

For calibration, we follow~\citet{vsliogeris2025full} and use $N{=}1{,}000$ examples sampled from MMLU~\citep{hendrycks2020measuring}. The calibration data is separate from the edit requests and is used only to estimate parameter sensitivity. Each example is processed individually, so the squared gradients correspond to per-example empirical Fisher estimates rather than gradients averaged over a batch.

\subsection{Theoretical Interpretation}
\label{sec:theory}

{In this section, we explain why the element-wise Fisher regularizer in Eqn.~\eqref{eq:goal} provides an effective regularization principle for model editing. Specifically, we relate it to the locality cost of an update $\Delta$, measured by the KL divergence between the pre-edit and post-edit predictive distributions on a reference set $\mathcal{D}_{\mathrm{ref}}$:
\begin{equation}
\mathcal{L}_{\mathrm{loc}}(\Delta)
\! = \!
\mathbb{E}_{x \sim \mathcal{D}_{\mathrm{ref}}}
\!\Big[
\mathrm{KL}\bigl(p_{\mathbf{W}}(\cdot \! \mid \! x)\,\|\,p_{\mathbf{W}+\Delta}(\cdot \! \mid \! x)\bigr)
\! \Big].
\label{eq:loc}
\end{equation}
To establish this connection, we consider a second-order expansion of $\mathcal{L}_{\mathrm{loc}}$ around $\Delta=0$. Under the regularity conditions in Appendix~\ref{app:regularity}, its value and gradient vanish at this point, and its Hessian equals the Fisher matrix $\mathbf{F}^{(S)}$ restricted to the edited parameters. Writing $u=\mathrm{vec}(\Delta)$, we obtain
\begin{equation}
\mathcal{L}_{\mathrm{loc}}(\Delta)
=
\tfrac{1}{2}\,u^\top
\mathbf{F}^{(S)}
u
+
o(\|u\|^2).
\end{equation}
The full $md\times md$ Fisher is intractable at LLM scale. Following~\citet{kirkpatrick2017overcoming}, we use a diagonal approximation, estimated in practice by the empirical Fisher in Eqn.~\eqref{eq:fisher_diag}:
\begin{equation}
\tfrac{1}{2}\,u^\top \mathbf{F}^{(S)}u
\;\approx\;
\tfrac{1}{2}\sum_{i,j} F_{ij}\,\Delta_{ij}^2,
\end{equation}
which recovers the form of the regularizer in Eqn.~\eqref{eq:goal}, with the constant absorbed into $\lambda$. FisherEdit therefore uses empirical parameter sensitivities to approximate the local cost of changing the pre-edit distribution, providing a tractable regularization principle for preserving model behavior. The full Taylor expansion and derivation are provided in Appendix~\ref{app:proof}.}

\section{Experiments}
\label{sec:exp}

We conduct comprehensive experiments to answer two core questions: (1) How does FisherEdit improve the overall performance of editing? (2) Does FisherEdit protect the general performance?

\begin{table}[t]
    \centering
    \small
    \setlength{\tabcolsep}{3pt}
    \renewcommand{\arraystretch}{1.08}
    \definecolor{fisherbg}{HTML}{E8F0FE}
    \definecolor{avgbg}{HTML}{F4F1E8}
    \caption{CounterFact results across three backbones under sequential editing. Best per column in \textbf{bold}; second best underlined.}
    \label{tab:main_results_counterfact}
    \begin{adjustbox}{max width=\textwidth}
    \begin{tabular}{l ccc >{\columncolor{avgbg}}c ccc >{\columncolor{avgbg}}c ccc >{\columncolor{avgbg}}c}
        \toprule
        & \multicolumn{4}{c}{\textbf{Llama-3-8B-Instruct}} & \multicolumn{4}{c}{\textbf{Qwen2.5-7B-Instruct}} & \multicolumn{4}{c}{\textbf{Olmo-3-7B-Instruct}} \\
        \cmidrule(lr){2-5} \cmidrule(lr){6-9} \cmidrule(lr){10-13}
        \textbf{Method} & \textbf{Eff} & \textbf{Gen} & \textbf{Spe} & \textbf{Avg} & \textbf{Eff} & \textbf{Gen} & \textbf{Spe} & \textbf{Avg} & \textbf{Eff} & \textbf{Gen} & \textbf{Spe} & \textbf{Avg} \\
        \midrule
        FT        & 65.02 & 58.05 & 41.63 & 54.90 & 73.23 & 62.76 & 40.52 & 58.84 & 95.17 & 70.87 & 56.12 & 74.05 \\
        ROME      & 33.84 & 36.51 & 68.16 & 46.17 & 37.33 & 40.92 & 65.52 & 47.92 & 36.10 & 39.12 & 69.32 & 48.18 \\
        MEMIT     & 48.00 & 47.64 & 52.52 & 49.39 & 64.20 & 62.41 & 55.88 & 60.83 & 71.38 & 69.53 & 55.45 & 65.45 \\
        MEND      & 48.41 & 48.41 & 51.05 & 49.29 & 1.13 & 0.67 & 1.15 & 0.98 & 38.97 & 43.13 & 62.18 & 48.09 \\
        GRACE     & 21.84 & 26.15 & \textbf{78.13} & 42.04 & 37.23 & 41.94 & 64.55 & 47.91 & 39.17 & 42.00 & 64.09 & 48.42 \\
        RECT      & 33.94 & 36.87 & 70.05 & 46.95 & 36.20 & 40.61 & 65.87 & 47.56 & 36.71 & 39.89 & 68.39 & 48.33 \\
        PRUNE     & 35.48 & 37.48 & 67.88 & 46.95 & 37.33 & 40.92 & 65.52 & 47.92 & 36.00 & 38.71 & 69.47 & 48.06 \\
        WISE      & 43.59 & 42.31 & 61.55 & 49.15 & 44.21 & 45.23 & 61.76 & 50.40 & 39.28 & 42.10 & 63.80 & 48.39 \\
        \midrule

        AnyEdit   & 52.00 & 47.33 & 51.72 & 50.35 & 69.53 & 48.05 & 56.34 & 57.97 & 63.48 & 51.28 & 48.29 & 54.35 \\
        RLEdit    & 24.20 & 27.43 & 75.63 & 42.42 & 35.79 & 38.20 & 66.42 & 46.80 & 22.25 & 29.94 & \textbf{75.63} & 42.61 \\
        UltraEdit & 52.41 & 53.12 & 46.77 & 50.77 & 60.61 & 42.05 & \textbf{69.57} & 57.41 & 43.38 & 45.33 & 57.57 & 48.76 \\
        AlphaEdit & \underline{99.79} & \underline{90.15} & 73.25 & \underline{87.73} & \textbf{100.00} & \underline{64.15} & 66.36 & \underline{76.84} & \textbf{99.48} & \underline{75.84} & 72.76 & \underline{82.69} \\
        \rowcolor{fisherbg}\textbf{FisherEdit (ours)} & \textbf{99.90} & \textbf{90.92} & \underline{75.68} & \textbf{88.83} & \underline{99.90} & \textbf{71.08} & \underline{66.66} & \textbf{79.21} & \underline{99.38} & \textbf{79.53} & \underline{73.33} & \textbf{84.08} \\
        \bottomrule
    \end{tabular}
    \end{adjustbox}
\end{table}

\begin{table}[t]
    \centering
    \small
    \setlength{\tabcolsep}{3pt}
    \renewcommand{\arraystretch}{1.08}
    \definecolor{fisherbg}{HTML}{E8F0FE}
    \definecolor{avgbg}{HTML}{F4F1E8}
    \caption{Main results across CounterFact, Ripple, and EVOKE on Olmo-3-7B-Instruct under sequential editing. Best per column in \textbf{bold}; second best underlined.}
    \label{tab:main_results_olmo}
    \begin{adjustbox}{max width=\textwidth}
    \begin{tabular}{l ccc >{\columncolor{avgbg}}c ccc >{\columncolor{avgbg}}c ccc >{\columncolor{avgbg}}c}
        \toprule
        & \multicolumn{4}{c}{\textbf{\textsc{CounterFact}}} & \multicolumn{4}{c}{\textbf{\textsc{Ripple}}} & \multicolumn{4}{c}{\textbf{\textsc{EVOKE}}} \\
        \cmidrule(lr){2-5} \cmidrule(lr){6-9} \cmidrule(lr){10-13}
        \textbf{Method} & \textbf{Eff} & \textbf{Gen} & \textbf{Spe} & \textbf{Avg} & \textbf{Eff} & \textbf{Gen} & \textbf{Spe} & \textbf{Avg} & \textbf{Eff} & \textbf{Gen} & \textbf{Spe} & \textbf{Avg} \\
        \midrule
        FT        & 95.17 & 70.87 & 56.12 & 74.05 & 76.10 & 59.87 & 63.79 & 66.59 & \underline{94.10} & 79.48 & 47.70 & 73.76 \\
        ROME      & 36.10 & 39.12 & 69.32 & 48.18 & 42.60 & 44.01 & 67.30 & 51.30 & 33.50 & 34.22 & 65.58 & 44.43 \\
        MEMIT     & 71.38 & 69.53 & 55.45 & 65.45 & 48.40 & 45.99 & 69.85 & 54.75 & 64.80 & 63.81 & 51.16 & 59.92 \\
        MEND      & 38.97 & 43.13 & 62.18 & 48.09 & 32.90 & 34.94 & 74.62 & 47.48 & 37.20 & 37.32 & 63.73 & 46.08 \\
        GRACE     & 39.17 & 42.00 & 64.09 & 48.42 & 50.20 & 48.39 & 60.89 & 53.16 & 35.40 & 35.80 & 63.52 & 44.91 \\
        RECT      & 36.71 & 39.89 & 68.39 & 48.33 & 41.00 & 41.91 & 68.93 & 50.61 & 34.00 & 34.42 & 65.66 & 44.69 \\
        PRUNE     & 36.00 & 38.71 & 69.47 & 48.06 & 43.80 & 43.93 & 67.17 & 51.63 & 33.30 & 34.21 & 65.59 & 44.37 \\
        WISE      & 39.28 & 42.10 & 63.80 & 48.39 & 56.20 & 56.71 & 55.06 & 55.99 & 36.00 & 35.96 & 62.77 & 44.91 \\
        \midrule
        AnyEdit   & 63.48 & 51.28 & 48.29 & 54.35 & 50.80 & 49.85 & \textbf{82.60} & 61.08 & 68.10 & 58.90 & 40.54 & 55.85 \\
        RLEdit    & 22.25 & 29.94 & \textbf{75.63} & 42.61 & 6.90 & 10.03 & \underline{82.55} & 33.16 & 17.10 & 18.83 & \textbf{78.51} & 38.15 \\
        UltraEdit & 43.38 & 45.33 & 57.57 & 48.76 & 48.00 & 49.04 & 57.03 & 51.36 & 48.90 & 48.88 & 51.19 & 49.66 \\
        AlphaEdit & \textbf{99.48} & \underline{75.84} & 72.76 & \underline{82.69} & \underline{96.70} & \textbf{75.10} & 33.39 & \underline{68.40} & \textbf{99.90} & \underline{92.47} & 70.14 & \underline{87.50} \\
        \rowcolor{fisherbg}\textbf{FisherEdit (ours)} & \underline{99.38} & \textbf{79.53} & \underline{73.33} & \textbf{84.08} & \textbf{97.30} & \underline{74.66} & 42.28 & \textbf{71.41} & \textbf{99.90} & \textbf{92.94} & \underline{71.08} & \textbf{87.97} \\
        \bottomrule
    \end{tabular}
    \end{adjustbox}
\end{table}

\subsection{Experiment Setup}

\noindent\textbf{Datasets and Models.}
We evaluate FisherEdit on three widely used model editing benchmarks: CounterFact~\citep{meng2022locating}, Ripple~\citep{cohen2024evaluating}, and EVOKE~\citep{zhang2025uncovering}. Experiments are conducted under a sequential editing setting with 1,000 edits, using three representative backbones: Llama-3-8B-Instruct~\citep{grattafiori2024llama3}, Qwen2.5-7B-Instruct~\citep{qwen2025qwen25technicalreport}, and Olmo-3-7B-Instruct~\citep{olmo2025olmo}.
To assess the impact on general capabilities, we further evaluate the edited models on HumanEval~\citep{chen2021evaluating}, ARC-Challenge/Easy~\citep{clark2018think}, IFEval~\citep{zhou2023instruction}, WinoGrande~\citep{sakaguchi2021winogrande} and GPQA~\citep{rein2023gpqa}, and report WikiText~\citep{merity2016pointer} perplexity after sequential editing on CounterFact.

\noindent\textbf{Baselines and Metrics.}
We compare FisherEdit with twelve baselines, including the learning-based and direct weight-editing methods discussed in Section~\ref{sec:related}, as well as FT~\citep{zhu2020modifying}, GRACE~\citep{hartvigsen2023aging}, WISE~\citep{wang2024wise}, AnyEdit~\citep{jiang2025anyedit}, and UltraEdit~\citep{gu2025ultraedit}.
For editing performance, we report efficacy (Eff), generalization (Gen), specificity (Spe), and their unweighted average (Avg).
For general performance, we report task accuracy and WikiText perplexity.

\subsection{Editing Performance}
\label{sec:main}

Table~\ref{tab:main_results_counterfact} reports CounterFact results across three backbones. FisherEdit achieves the highest Avg in all three settings, with gains over AlphaEdit of $+1.10$ on Llama, $+2.37$ on Qwen, and $+1.39$ on Olmo. Since Efficacy is already close to saturated for both AlphaEdit and FisherEdit, the main differences appear in Gen and Spe. This indicates that FisherEdit improves the quality of the update primarily by better balancing paraphrase transfer and neighborhood preservation.

Table~\ref{tab:main_results_olmo} further evaluates whether the same trend holds across datasets on a fixed backbone. On Olmo-3-7B-Instruct, FisherEdit obtains the best Avg on CounterFact, Ripple, and EVOKE. The improvements over AlphaEdit are consistent: $+1.39$, $+3.01$, and $+0.47$ points, respectively. These results support the claim that Fisher-weighted regularization provides a robust preservation mechanism across both model and benchmark variations.

{We also evaluate FisherEdit's computational efficiency by measuring
wall-clock time and peak GPU memory across the editing pipeline, with a
detailed comparison to AlphaEdit in Appendix~\ref{app:cost}. The results show
that FisherEdit substantially reduces preprocessing time and overall runtime
including preprocessing, while using nearly the same peak GPU memory.}

\subsection{General Performance}
\label{sec:general}

\begin{figure}[t]
    \centering
    \includegraphics[width=\linewidth]{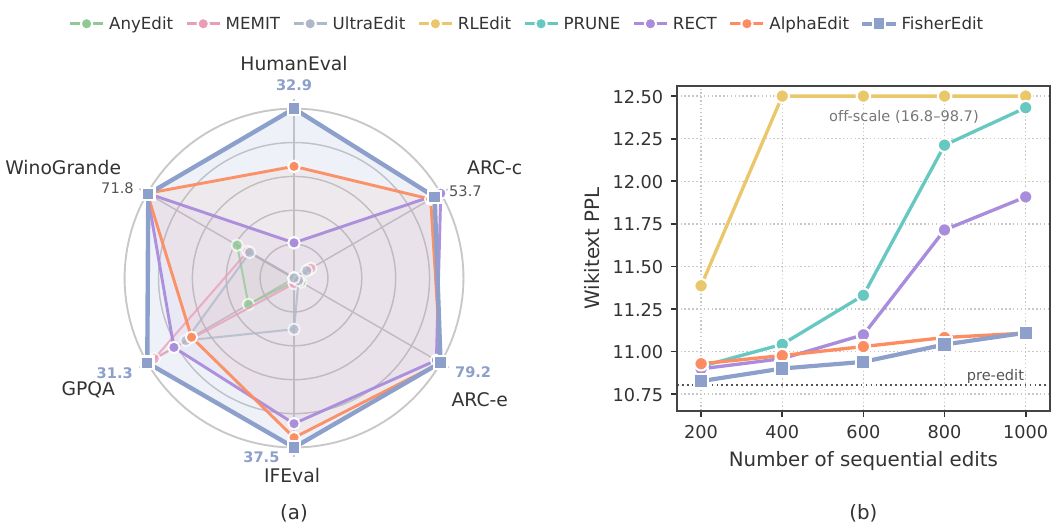}
    \caption{(a) Downstream task scores of Llama-3-8B-Instruct on HumanEval, ARC-c, ARC-e, IFEval, WinoGrande and GPQA. (b) Wikitext perplexity of Llama-3-8B-Instruct after applying $N\in\{200,400,600,800,1000\}$ CounterFact edits.}
    \label{fig:downstream_ppl}
\end{figure}

\noindent\textbf{General domains.}
Preserving general capabilities beyond the edited facts is crucial, yet is often overlooked by prior editing methods.
From Figure~\ref{fig:downstream_ppl}(a),
FisherEdit maintains stable performance across all six downstream tasks, covering code generation, science, commonsense reasoning, instruction following, and challenging knowledge-intensive question answering.
In contrast, AlphaEdit and RECT exhibit noticeable drops on HumanEval and GPQA.
These results suggest that the Fisher-diagonal regularizer helps mitigate collateral damage while still enabling effective edits.

\noindent\textbf{Language modeling.}
We further evaluate the impact of editing on general language modeling by reporting WikiText perplexity after sequentially editing $N\in\{200,400,600,800,1000\}$ CounterFact samples.
As shown in Figure~\ref{fig:downstream_ppl}(b), FisherEdit better preserves general text modeling ability than competing methods during sequential editing.
Its WikiText perplexity remains consistently low across all edit counts, increasing only mildly from $10.826$ after 200 edits to $11.110$ after 1,000 edits.
By contrast, AlphaEdit, RECT, and PRUNE show larger increases. These results indicate that FisherEdit maintains the most stable general-language perplexity throughout the editing process.

\subsection{Ablation Study}
\label{sec:ablation}

\begingroup
\addtolength{\columnsep}{4pt}
\noindent
\begin{wraptable}[12]{r}{0.46\linewidth}
    \vspace{-16pt}
    \centering
    \small
    \setlength{\tabcolsep}{4.5pt}
    \caption{Fisher-matrix ablation on CounterFact with Llama-3-8B-Instruct under sequential editing. Fisher diagonal $\mathbf{F}$ is replaced with an all-ones matrix (\emph{Uniform}) or with i.i.d.\ uniform random values (\emph{Random}).}
    \label{tab:fisher_ablation}
    \vspace{2pt}
    \begin{tabular}{lcccc}
        \toprule
        & \textbf{Eff} & \textbf{Gen} & \textbf{Spe} & \textbf{Avg} \\
        \midrule
        Fisher  & \textbf{99.58} & 91.28 & \textbf{75.13} & \textbf{88.66} \\
        Uniform & 99.28 & \textbf{91.48} & 72.22 & 87.66 \\
        Random  & 95.89 & 85.74 & 69.94 & 83.86 \\
        \bottomrule
    \end{tabular}
    \vspace{-8pt}
\end{wraptable}
\textbf{Fisher Matrix.} To examine whether the Fisher diagonal captures model-specific information, we replace $\mathbf{F}$ in Eq.~\eqref{eq:goal} with an all-ones matrix (\emph{Uniform}), which reduces the objective to $\ell_2$ regularization, or i.i.d.\ samples from $\mathrm{Uniform}[0,1]$ (\emph{Random}), while keeping all other hyperparameters fixed. From Table~\ref{tab:fisher_ablation}, replacing $\mathbf{F}$ with \emph{Uniform} largely preserves rewrite and paraphrase performance, which are mainly controlled by $\lambda$ as discussed in \S\ref{sec:rethinking}, but substantially weakens neighborhood locality.
Meanwhile, \emph{Random} degrades all three metrics.
These results indicate that FisherEdit benefits from the structured information encoded in the Fisher diagonal, rather than from generic non-uniform reweighting.
\par\endgroup

\noindent\textbf{Top-Fisher masking.}
{The Fisher diagonal is heavy-tailed and model-dependent: in layer~4, its largest $1\%$ of cached entries carry $16.8\%$, $52.6\%$, and $30.4\%$ of the total mass for Llama-3-8B-Instruct, Qwen2.5-7B-Instruct, and Olmo-3-7B-Instruct, respectively (a uniform matrix would carry $1\%$). To test whether the high-Fisher coordinates are important for preservation, we rerun the Llama-3-8B-Instruct/CounterFact setting with 1,000 sequential edits while selectively removing or retaining the highest-Fisher coordinates. For \emph{mask-top}, entries at the top $q$ percent are replaced by that layer's median; for \emph{keep-top}, only those entries are retained and all others are replaced by the median. As shown in Table~\ref{tab:top_fisher_masking_v2}, masking the top $1\%$ decreases Specificity by $1.65$ points, eliminating approximately $48\%$ of the Full-Fisher--Uniform Specificity gap and $63\%$ of the Average gap. Conversely, keeping only the top $10\%$ recovers approximately $60\%$ of the Specificity advantage and $70\%$ of the Average advantage, while the full diagonal remains strongest.  These results provide direct evidence that the proposed Fisher diagonal plays a fundamental role in FisherEdit's gains, beyond uniform $\ell_2$ regularization alone.}

\begin{table}[!hbp]
    \centering
    \small
    \setlength{\tabcolsep}{3.5pt}
    \begin{minipage}[t]{0.485\linewidth}
        \centering
        \caption{Effect of masking or retaining high-Fisher coordinates.}
        \label{tab:top_fisher_masking_v2}
        \vspace{3pt}
        \begin{tabular}{lcccc}
            \toprule
            \textbf{} & \textbf{Eff} & \textbf{Gen} & \textbf{Spe} & \textbf{Avg} \\
            \midrule
            Full Fisher       & \textbf{99.90} & 90.92 & \textbf{75.68} & \textbf{88.83} \\
            Mask top $1\%$    & 99.49 & 90.87 & 74.03 & 88.13 \\
            Mask top $10\%$   & 99.38 & 91.03 & 73.50 & 87.97 \\
            Mask top $30\%$   & 99.49 & 89.54 & 74.33 & 87.79 \\
            Keep top $10\%$   & 99.69 & \textbf{91.49} & 74.29 & 88.49 \\
            Uniform           & 99.90 & 91.02 & 72.22 & 87.71 \\
            \bottomrule
        \end{tabular}
    \end{minipage}\hfill
    \begin{minipage}[t]{0.485\linewidth}
        \centering
        \caption{Ablation on the number of MMLU examples $n$ used to estimate the Fisher diagonal.}
        \label{tab:fisher_sample_size_ablation}
        \vspace{3pt}
        \begin{tabular}{lcccc}
            \toprule
            \textbf{$n$} & \textbf{Eff} & \textbf{Gen} & \textbf{Spe} & \textbf{Avg} \\
            \midrule
            50   & 99.79 & 89.84 & 74.18 & 87.94 \\
            100  & 99.69 & \textbf{92.15} & 73.48 & 88.44 \\
            500  & 99.79 & 91.12 & 75.10 & \textbf{88.67} \\
            1000 & 99.58 & 91.28 & \textbf{75.13} & 88.66 \\
            2000 & \textbf{99.79} & 90.76 & 74.66 & 88.40 \\
            \bottomrule
        \end{tabular}
    \end{minipage}
\end{table}

\noindent\textbf{Number of Calibration Examples.} We vary the number of examples $n$ used to estimate the Fisher diagonal while keeping the editing configuration fixed.
As shown in Table~\ref{tab:fisher_sample_size_ablation}, FisherEdit is largely insensitive to the estimation sample size: the average score remains within a narrow range, from $87.94$ to $88.67$, and the neighborhood score changes only mildly.
Even with only $n{=}50$ examples, the Fisher constraint retains most of its benefit, suggesting that the regularizer does not require a large reference set to provide effective protection.

\section{Conclusion}
\label{sec:conclusion}
{In this work, we addressed parameter interference and capability degradation in sequential knowledge editing with FisherEdit, a Fisher-guided framework that replaces coarse activation-space projection with importance-aware parameter regularization and computes updates through an efficient row-wise closed-form solver. Theoretical analysis and empirical results show that FisherEdit reduces distributional drift and achieves state-of-the-art editing performance while better preserving general capabilities across backbones and benchmarks.}

\subsection*{AI Use Statement}
{AI assistants were used only for language polishing, \LaTeX{} checking,
and coding/debugging assistance. All scientific claims, experiments, analyses,
and final writing were verified and controlled by the authors.}

\subsection*{Reproducibility Statement}
{The editing algorithm, datasets, evaluation metrics, and hyperparameters are
specified in Appendices~\ref{app:fisheredit_flow} and~\ref{app:details}.
Appendices~\ref{app:row_solution_derivation} and~\ref{app:proof} provide the
derivations, and Appendix~\ref{app:seeds} reports results across random seeds.}

\bibliographystyle{iclr2027_conference}
\bibliography{custom}

\clearpage
\appendix

\section{FisherEdit Algorithm Flow}
\label{app:fisheredit_flow}

\begin{algorithm}[H]
\caption{FisherEdit sequential editing}
\label{alg:fisheredit}
\begin{algorithmic}[1]
\REQUIRE pretrained model $\mathbf{W}$, edited layers $\mathcal{L}$, edit stream $\{\mathcal{B}_t\}_{t=1}^T$, diagonal Fisher $\mathbf{F}$, regularization strength $\lambda$
\STATE Initialize historical key memory $\mathbf{K}_p^{(\ell)} \leftarrow \varnothing$ for each layer $\ell \in \mathcal{L}$
\FOR{$t = 1$ to $T$}
    \STATE Format the current batch prompts and compute the target representation $z^\star$ for each edit request
    \FOR{each layer $\ell \in \mathcal{L}$}
        \STATE Trace the current lookup keys $\mathbf{K}_1^{(\ell)}$ and compute the remaining residual $\mathbf{E}^{(\ell)}$
        \STATE Retrieve historical keys $\mathbf{K}_p^{(\ell)}$ from memory
        \STATE Form $\mathbf{U}^{(\ell)} = [\mathbf{K}_1^{(\ell)}, \mathbf{K}_p^{(\ell)}]$ and $\mathbf{E}'^{(\ell)} = [\mathbf{E}^{(\ell)}, \mathbf{0}]$
        \STATE Solve the Fisher-weighted row-wise update $\Delta^{(\ell)}$ by Eqn.~\eqref{eq:efficient_row_solution}
        \STATE Update the layer weights: $\mathbf{W}^{(\ell)} \leftarrow \mathbf{W}^{(\ell)} + \Delta^{(\ell)}$
        \STATE Append the current batch keys to memory: $\mathbf{K}_p^{(\ell)} \leftarrow [\mathbf{K}_p^{(\ell)}, \mathbf{K}_1^{(\ell)}]$
    \ENDFOR
\ENDFOR
\STATE \textbf{return} the edited model
\end{algorithmic}
\end{algorithm}

\section{Experimental Details}
\label{app:details}

\subsection{Evaluation Metrics}
\label{app:metrics}

For an edit request $(s_i,o_i)\to o_i^*$, where $s_i$ is the subject, $o_i$ is the original object, and $o_i^*$ is the edited target object, let $q_i^{\mathrm{edit}}$ denote the rewrite prompt used to state the edited fact, $\mathcal{Q}_i^{\mathrm{para}}$ the set of paraphrase prompts that express the same edited fact, and $\mathcal{Q}_i^{\mathrm{loc}}$ the set of neighborhood prompts used to test preserved knowledge. After applying the edit sequence, we evaluate all metrics with the post-edit model $p_{W+\Delta}$.

For a prompt $q$ and a candidate answer $a=(a_1,\ldots,a_T)$, define the average target negative log-likelihood
\begin{equation}
\ell_{\Delta}(a\mid q)
= -\frac{1}{T}\sum_{t=1}^{T}
\log p_{\mathbf{W}+\Delta}(a_t \mid q, a_{<t}).
\label{eq:metric-nll}
\end{equation}
Lower $\ell_{\Delta}(a\mid q)$ means that the edited model assigns higher probability to answer $a$ under prompt $q$.

\paragraph{Efficacy.}
Efficacy measures whether the edited model recalls the new target on the original rewrite prompt. A request is counted as successful when the new answer has lower average NLL than the original answer:
\begin{equation}
\begin{aligned}
e_i &=
\mathbf{1}\!\left[
\ell_{\Delta}(o_i^*\mid q_i^{\mathrm{edit}})
<
\ell_{\Delta}(o_i\mid q_i^{\mathrm{edit}})
\right],\\
\mathrm{Eff}
&= \frac{1}{n}\sum_{i=1}^{n} e_i .
\end{aligned}
\label{eq:metric-efficacy}
\end{equation}

\paragraph{Generalization.}
Generalization measures whether the same update transfers to paraphrased prompts. For each request, we average the same new-versus-original comparison over its paraphrase set:
\begin{equation}
\begin{aligned}
g_i(q) &=
\mathbf{1}\!\left[
\ell_{\Delta}(o_i^*\mid q)
<
\ell_{\Delta}(o_i\mid q)
\right],\\
\mathrm{Gen}
&= \frac{1}{n}\sum_{i=1}^{n}
\frac{1}{|\mathcal{Q}_i^{\mathrm{para}}|}
\sum_{q\in\mathcal{Q}_i^{\mathrm{para}}}
g_i(q).
\end{aligned}
\label{eq:metric-generalization}
\end{equation}

\paragraph{Specificity.}
Specificity measures whether unrelated or neighboring facts are preserved. Let $o_i^{\mathrm{loc}}(q)$ be the correct preserved answer for a neighborhood prompt $q$. A neighborhood prompt is counted as successful when the preserved answer remains more likely than the edit target:
\begin{equation}
\begin{aligned}
s_i(q) &=
\mathbf{1}\!\left[
\ell_{\Delta}(o_i^{\mathrm{loc}}(q)\mid q)
<
\ell_{\Delta}(o_i^*\mid q)
\right],\\
\mathrm{Spe}
&= \frac{1}{n}\sum_{i=1}^{n}
\frac{1}{|\mathcal{Q}_i^{\mathrm{loc}}|}
\sum_{q\in\mathcal{Q}_i^{\mathrm{loc}}}
s_i(q).
\end{aligned}
\label{eq:metric-specificity}
\end{equation}

\subsection{Dataset Details}
\label{app:dataset_details}

\paragraph{Main editing datasets.}
CounterFact~\citep{meng2022locating} is a factual knowledge editing benchmark in which each request rewrites a subject--relation association from an original object to a counterfactual target object. Each record also provides paraphrase prompts for testing whether the edited fact generalizes beyond the original wording, as well as neighborhood prompts for testing whether unrelated facts are preserved. We use the AKEW~\citep{wu2024akew} version of CounterFact and remove records whose aligned AKEW annotations are unavailable, leaving 975 valid edit requests. RippleEdits~\citep{cohen2024evaluating} evaluates whether an edit has appropriate ripple effects on related queries while preserving other relation-specific knowledge about the edited subject. In our experiments, we use the \texttt{random} subset of RippleEdits and evaluate the first 1,000 records. EVOKE~\citep{zhang2025uncovering} is designed to evaluate overfitting in model editing by combining direct rewrite prompts with augmented paraphrase prompts and locality prompts. We use the \texttt{main} subset of EVOKE and evaluate the first 1,000 records. All main editing results are reported under sequential editing with 100 edit requests per batch. The edited model is not reset between batches; after processing the full edit stream, we evaluate the final edited model on all requests.

\paragraph{Downstream datasets.}
To measure whether editing damages general capabilities beyond the edited facts, we evaluate the edited models on five downstream benchmarks. HumanEval~\citep{chen2021evaluating} evaluates code generation by asking models to complete Python programming problems, and we report pass@1. ARC-Challenge and ARC-Easy~\citep{clark2018think} evaluate multiple-choice science question answering, with ARC-Challenge containing the harder subset and ARC-Easy containing the easier subset; we report normalized accuracy. IFEval~\citep{zhou2023instruction} evaluates instruction-following behavior by checking whether model outputs satisfy verifiable prompt-level constraints, and we report strict prompt accuracy. WinoGrande~\citep{sakaguchi2021winogrande} evaluates commonsense reasoning through adversarial Winograd-style pronoun resolution examples, and we report accuracy. GPQA~\citep{rein2023gpqa} evaluates difficult graduate-level, Google-proof multiple-choice questions; we use the Diamond zero-shot setting and report normalized accuracy. We additionally report WikiText~\citep{merity2016pointer} perplexity to measure general language-modeling quality after sequentially applying CounterFact edits.

\subsection{Hyperparameter Configurations}
\label{app:hyperparameter_configurations}

All experiments use instruction-tuned backbones: Llama-3-8B-Instruct, Qwen2.5-7B-Instruct, and Olmo-3-7B-Instruct. Models are loaded with HuggingFace \texttt{AutoModelForCausalLM} and evaluated with their corresponding tokenizers. Because the backbones are instruction-tuned, both editing prompts and evaluation prompts are serialized with each tokenizer's chat template. We run the entire editing pipeline in fp32, including offline precomputation, and use bf16 for downstream inference. We edit the MLP down-projection matrices \texttt{model.layers.\{\}.mlp.down\_proj} at layers 4--8, use the last subject token as the fact lookup token, and optimize target values for 25 gradient steps with learning rate 0.1 and weight decay 0.5. The target loss layer to calculate $\mathbf{V}$ is 31 for Llama-3-8B-Instruct and Olmo-3-7B-Instruct, and 27 for Qwen2.5-7B-Instruct. We set $\lambda=10^{-3}$ across all benchmarks and backbones. For Fisher diagonal estimation, we estimates the empirical Fisher diagonal once from 1,000 MMLU examples, computing per-example squared cross-entropy gradients only for the edited MLP down-projection weights. The resulting Fisher tensors are cached and reused throughout the sequential edit stream.

\section{Additional Experimental Results}
\label{app:additional}

\subsection{Seed Variance and Significance Testing}
\label{app:seeds}

{Table~\ref{tab:main_results_counterfact} reports single runs under the
benchmark's native edit order. To quantify run-to-run variability, we repeat
the experiment with three random seeds for both
FisherEdit and AlphaEdit. Each seed reshuffles the edit stream, i.e.\ the
composition and ordering of the sequential batches, and reseeds all stochastic
components including the target-value optimization. Both methods see exactly
the same cases under matched seeds, so the comparison is paired.}

\begin{table}[h]
    \centering
    \small
    \caption{{Three-seed comparison on CounterFact with
    Llama-3-8B-Instruct. Mean $\pm$ standard deviation over seeds; each seed
    reshuffles the sequential edit stream.}}
    \label{tab:seed_variance}
    \begin{tabular}{lcccc}
        \toprule
        \textbf{Method} & \textbf{Eff} & \textbf{Gen} & \textbf{Spe} & \textbf{Avg} \\
        \midrule
        AlphaEdit  & 99.69 \small{$\pm$ 0.10} & 90.17 \small{$\pm$ 0.75} & 73.75 \small{$\pm$ 0.30} & 87.87 \small{$\pm$ 0.28} \\
        FisherEdit & \textbf{99.79} \small{$\pm$ 0.21} & \textbf{92.68} \small{$\pm$ 0.70} & \textbf{73.91} \small{$\pm$ 0.16} & \textbf{88.80} \small{$\pm$ 0.28} \\
        \bottomrule
    \end{tabular}
\end{table}

{The largest gap is in Generalization, where FisherEdit leads by $2.51$
points on average. A two-sided Welch's $t$-test on the three seed-level
Generalization scores gives $p=0.013$. The Average advantage is also
consistent rather than seed-specific: FisherEdit is higher in all three paired
runs, by $+0.76$ to $+1.03$ points (mean $+0.93$). Efficacy is near saturation
for both methods and Specificity is comparable, so the seed-level picture
matches the single-run trend in \S\ref{sec:main}.}

\subsection{Results on Mistral-7B-Instruct}
\label{app:mistral}

{To check that the conclusion is not tied to the three backbones of
\S\ref{sec:exp}, we extend the evaluation on Mistral-7B-Instruct-v0.3 on
both CounterFact and Ripple, against the six  baselines from
Tables~\ref{tab:main_results_counterfact} and~\ref{tab:main_results_olmo}.}

\begin{table}[h]
    \centering
    \small
    \setlength{\tabcolsep}{4pt}
    \caption{{Sequential editing on Mistral-7B-Instruct-v0.3.}}
    \label{tab:mistral}
    \begin{tabular}{l cccc cccc}
        \toprule
        & \multicolumn{4}{c}{\textbf{\textsc{CounterFact}}} & \multicolumn{4}{c}{\textbf{\textsc{Ripple}}} \\
        \cmidrule(lr){2-5} \cmidrule(lr){6-9}
        \textbf{Method} & \textbf{Eff} & \textbf{Gen} & \textbf{Spe} & \textbf{Avg} & \textbf{Eff} & \textbf{Gen} & \textbf{Spe} & \textbf{Avg} \\
        \midrule
        MEMIT     & 50.26 & 50.26 & 49.74 & 50.09 & 53.80 & 53.58 & 49.49 & 52.29 \\
        RECT      & 29.85 & 31.54 & 75.46 & 45.61 & 31.50 & 32.38 & \textbf{76.15} & 46.68 \\
        PRUNE     & 32.41 & 32.36 & 71.19 & 45.32 & 32.50 & 33.81 & \underline{73.24} & 46.52 \\
        RLEdit    & 48.00 & 48.26 & 52.91 & 49.72 & 49.30 & 47.19 & 55.62 & 50.70 \\
        UltraEdit & 54.26 & 54.15 & 45.83 & 51.41 & 67.60 & 69.38 & 34.58 & 57.19 \\
        AlphaEdit & \textbf{99.90} & 89.18 & \underline{76.91} & \underline{88.66} & \textbf{96.40} & \textbf{76.36} & 45.67 & \underline{72.81} \\
        \textbf{FisherEdit (ours)} & \textbf{99.90} & \textbf{90.10} & \textbf{77.12} & \textbf{89.04} & \underline{95.70} & \underline{75.99} & 51.12 & \textbf{74.27} \\
        \bottomrule
    \end{tabular}
\end{table}

{FisherEdit attains the best Average on both benchmarks. On CounterFact
it matches AlphaEdit's saturated Efficacy while improving Generalization and
Specificity ($88.66\rightarrow89.04$ Avg). On Ripple it trades a small amount
of rewrite and paraphrase accuracy for a $5.45$-point Specificity gain, giving
$+1.46$ Avg. Overall, it extends the backbone-level
consistency of our conclusion to four instruction-tuned model families.}

\subsection{AlphaEdit under Matched Regularization Strength}
\label{app:alpha_l2}

{To verify that FisherEdit's gain does not come merely from using a
weaker regularization strength than AlphaEdit's published $\ell_2=1$, we sweep
AlphaEdit over $\ell_2\in\{1,10^{-1},10^{-2},10^{-3}\}$ on all three
backbones in Table~\ref{tab:alpha_l2_sweep}.}

{Three orders of magnitude in $\ell_2$ move AlphaEdit's Average by at
most $0.81$ points, and the best setting is model-dependent. FisherEdit stays
ahead of every tested coefficient, by $+0.29$, $+1.89$, and $+0.98$ Average
points on Llama, Qwen, and Olmo. Tuning a scalar strength therefore does not
reproduce what the Fisher diagonal provides.}

\begin{table}[h]
\centering
\small
\setlength{\tabcolsep}{4.5pt}
\renewcommand{\arraystretch}{1.15}
\caption{{AlphaEdit with swept $\ell_2$ strength versus FisherEdit on
CounterFact.}}
\label{tab:alpha_l2_sweep}
\begin{adjustbox}{max width=\textwidth}
\begin{tabular}{clccccc}
\toprule
\textbf{Backbone} & \textbf{Method} & \textbf{$\ell_2$}
& \textbf{Eff} & \textbf{Gen} & \textbf{Spe} & \textbf{Avg} \\
\midrule

\multirow{5}{*}{Llama}
& \multirow{4}{*}{AlphaEdit} & $1$         & 99.79 & 90.15 & 73.25 & 87.73 \\
&                            & $10^{-1}$   & 99.79 & 91.44 & 74.17 & 88.47 \\
&                            & $10^{-2}$   & 99.79 & 91.18 & 74.56 & 88.51 \\
&                            & $10^{-3}$   & 99.90 & 90.77 & 74.95 & 88.54 \\
& \textbf{FisherEdit}        & $10^{-3}$   & 99.90 & 90.92 & 75.68 & \textbf{88.83} \\

\midrule

\multirow{5}{*}{Qwen}
& \multirow{4}{*}{AlphaEdit} & $1$         & 100.00 & 64.15 & 66.36 & 76.84 \\
&                            & $10^{-1}$   & 99.79  & 65.79 & 66.38 & 77.32 \\
&                            & $10^{-2}$   & 100.00 & 65.08 & 66.19 & 77.09 \\
&                            & $10^{-3}$   & 99.28  & 65.44 & 62.55 & 75.76 \\
& \textbf{FisherEdit}        & $10^{-3}$   & 99.90 & 71.08 & 66.66 & \textbf{79.21} \\

\midrule

\multirow{5}{*}{Olmo}
& \multirow{4}{*}{AlphaEdit} & $1$         & 99.48 & 75.84 & 72.76 & 82.69 \\
&                            & $10^{-1}$   & 99.59 & 76.46 & 72.87 & 82.97 \\
&                            & $10^{-2}$   & 99.59 & 76.41 & 72.95 & 82.98 \\
&                            & $10^{-3}$   & 99.59 & 76.62 & 73.08 & 83.10 \\
& \textbf{FisherEdit}        & $10^{-3}$   & 99.38 & 79.53 & 73.33 & \textbf{84.08} \\

\bottomrule
\end{tabular}
\end{adjustbox}
\end{table}

\subsection{Fisher Drift and Online Fisher Updating}
\label{app:online_fisher}

{FisherEdit estimates the Fisher diagonal once before editing and reuses
it throughout the edit stream. Because the edited weights change during
sequential editing, parameter importance could in principle drift, and
re-estimating the Fisher online might then help. We examine both questions
under the Llama-3-8B-Instruct / CounterFact setting of
Table~\ref{tab:main_results_counterfact} with 1,000 sequential edits.}

\paragraph{Parameter-importance drift.}
{We save the model after 1,000 sequential edits and re-estimate its Fisher
diagonal on the same held-out calibration set used before editing. We then
compare the pre- and post-editing diagonals at each edited layer using
log-Pearson and Spearman correlation, cosine similarity, and the overlap of
the top-$1\%$ and top-$10\%$ coordinate sets.}

\begin{table}[h]
    \centering
    \small
    \setlength{\tabcolsep}{5pt}
    \caption{{Agreement between the pre- and post-editing Fisher diagonals
    after 1,000 sequential edits, per edited layer.}}
    \label{tab:fisher_drift}
    \begin{tabular}{lccccc}
        \toprule
        \textbf{Layer} & \textbf{Pearson (log)} & \textbf{Spearman} & \textbf{Cosine} & \textbf{Top-1\% overlap} & \textbf{Top-10\% overlap} \\
        \midrule
        4 & 0.878 & 0.874 & 0.824 & 0.61 & 0.73 \\
        5 & 0.871 & 0.859 & 0.947 & 0.61 & 0.72 \\
        6 & 0.883 & 0.864 & 0.979 & 0.65 & 0.72 \\
        7 & 0.911 & 0.893 & 0.981 & 0.71 & 0.75 \\
        8 & 0.913 & 0.897 & 0.919 & 0.71 & 0.75 \\
        \bottomrule
    \end{tabular}
\end{table}

{As Table~\ref{tab:fisher_drift} shows, the importance structure remains
strongly aligned after 1,000 edits: Spearman correlation is $0.86$--$0.90$,
log-Pearson correlation is $0.87$--$0.91$, cosine similarity is
$0.82$--$0.98$, and $72$--$75\%$ of the highest-Fisher $10\%$ of parameters
remain in that set. Both the ranking and the membership of the high-Fisher
parameter sets are therefore largely stable under sequential editing with
FisherEdit.}

\paragraph{Online Fisher variant.}
{To test whether online re-estimation nevertheless helps, we implement a
variant that updates the Fisher after every edit batch. Let $\mathbf{F}_t$ be
the Fisher used for batch $t$. After applying the edits of batch $t$, we run
the updated model on the just-edited prompt/target pairs, compute per-example
squared cross-entropy gradients for the edited down-projection weights, and
average them into a local diagonal $\mathbf{F}_{\mathrm{local},t}$. The Fisher
for the next batch is
\begin{equation}
\mathbf{F}_{t+1}=\gamma\,\mathbf{F}_t+(1-\gamma)\,\mathbf{F}_{\mathrm{local},t},
\label{eq:online_fisher}
\end{equation}
where $\gamma\in[0,1]$ controls how quickly the estimate adapts. Setting
$\gamma=1$ exactly recovers the fixed Fisher used in the paper, while smaller
$\gamma$ gives more weight to the newly computed local Fisher. All other
editing settings are unchanged.}

\begin{table}[h]
    \centering
    \small
    \setlength{\tabcolsep}{5pt}
    \caption{{Fixed versus online Fisher on CounterFact with
    Llama-3-8B-Instruct. Extra time is the additional wall-clock cost per
    100-edit batch relative to the fixed Fisher.}}
    \label{tab:online_fisher}
    \begin{tabular}{lccccc}
        \toprule
        \textbf{Fisher} & \textbf{Eff} & \textbf{Gen} & \textbf{Spe} & \textbf{Avg} & \textbf{Extra time / batch} \\
        \midrule
        Fixed ($\gamma=1$)     & \textbf{99.90} & \textbf{90.92} & \textbf{75.68} & \textbf{88.83} & 0 \\
        Online ($\gamma=0.99$) & 99.59 & 90.21 & 75.14 & 88.31 & $\sim$95 s \\
        Online ($\gamma=0.9$)  & 99.79 & 90.56 & 74.87 & 88.41 & $\sim$97 s \\
        \bottomrule
    \end{tabular}
\end{table}

{Table~\ref{tab:online_fisher} shows that online updating provides no
improvement: both online variants fall slightly below the fixed Fisher on all
four metrics, with Average decreasing by $0.42$--$0.52$ points. They also
require an additional forward and backward pass over every edit batch, adding
roughly $95$--$97$ seconds per 100 edits, about $45\%$ of the per-batch editing
time reported in Table~\ref{tab:cost}. Together with the limited drift in
Table~\ref{tab:fisher_drift}, these results support the fixed Fisher as the
more effective and more efficient choice.}

\subsection{Computational Cost}
\label{app:cost}

{We instrument every phase of the pipeline and report wall-clock time and
peak allocated GPU memory on a single RTX 6000 Ada, under the same setting of
Table~\ref{tab:main_results_counterfact}.}

\begin{table}[h]
    \centering
    \small
    \setlength{\tabcolsep}{5pt}
    \caption{{Phase-wise computational cost relative to AlphaEdit.}}
    \label{tab:cost}
    \begin{tabular}{lccc}
        \toprule
        \textbf{Phase} & \textbf{AlphaEdit} & \textbf{FisherEdit} & \textbf{Diff.} \\
        \midrule
        \multicolumn{4}{l}{\textit{Wall-clock time}} \\
        Preprocessing (min) 
            & 648.0 & \textbf{17.7} & $-97.3\%$ \\
        Editing / 100 edits (s) 
            & \textbf{195.7} & 213.0 & $+8.9\%$ \\
        Evaluation (s) 
            & 328.4 & \textbf{318.7} & $-3.0\%$ \\
        End-to-end (min) 
            & 680.4 & \textbf{53.4} & $-92.1\%$ \\
        \midrule
        \multicolumn{4}{l}{\textit{Peak memory}} \\
        Peak GPU memory (GiB) 
            & 35.19 & \textbf{34.94} & $-0.7\%$ \\
        \bottomrule
    \end{tabular}
\end{table}

{FisherEdit substantially lowers the cost of constructing the
preservation constraint. AlphaEdit requires second-moment computation over
100k Wikipedia samples and layerwise SVD, taking 10.80 hours in our
experiment. FisherEdit estimates a diagonal Fisher matrix once from 1,000
MMLU examples and completes preprocessing in 17.70 minutes. Thus, FisherEdit
reduces preprocessing time by $36.6\times$ and end-to-end runtime by
$12.8\times$ for 1,000 sequential edits, without introducing inference-time
overhead.}

{The trade-off is an $8.9\%$ increase in per-batch editing time
($195.7$ versus $213.0$ seconds). This overhead is caused by the
Fisher-weighted row-wise solve, while the Woodbury formulation keeps the
computation efficient and peak GPU memory nearly unchanged.}

\section{Hidden-Space Drift Analysis}
\label{app:hidden_space_analysis}

\begin{figure}[t]
    \centering
    \includegraphics[width=\textwidth]{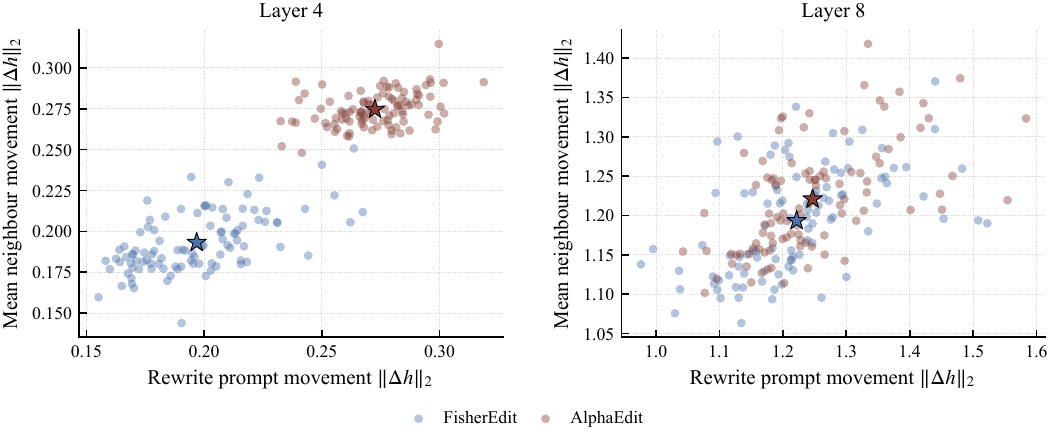}
    \caption{Hidden-state movement of rewrite and neighborhood prompts at layers 4 and 8.}
    \label{fig:hidden_tradeoff}
\end{figure}

To better understand how FisherEdit changes internal representations, we visualize the hidden-state movement induced by FisherEdit and AlphaEdit. We use the EVOKE sequential editing run on Olmo-3-7B-Instruct and select the first and last edited layers, layers 4 and 8. For each edit request, we record the last-token hidden state of the rewrite prompt and ten neighborhood prompts before and after editing. Each point in Figure~\ref{fig:hidden_tradeoff} corresponds to one edit request: the x-axis measures the $\ell_2$ movement of the rewrite prompt, while the y-axis measures the average $\ell_2$ movement of its neighborhood prompts. Lower neighborhood movement indicates less collateral change in nearby but unrelated contexts.

At layer 4, FisherEdit produces a clearly smaller  movement than AlphaEdit, indicating that the Fisher-weighted constraint already limits unintended representation drift at the first edited layer. At layer 8, the advantage becomes less pronounced but remains visible. This attenuation is expected, since later edited layers receive representations that have already been affected by preceding edited layers, making the hidden-space comparison noisier. Overall, the case study suggests that FisherEdit better controls hidden representation drift, especially at the early stage of the edited layer stack.

\section{Weight-Space View of Regularization}
\label{app:delta_norm}

\begin{figure}[h]
    \centering
    \includegraphics[width=0.5\linewidth]{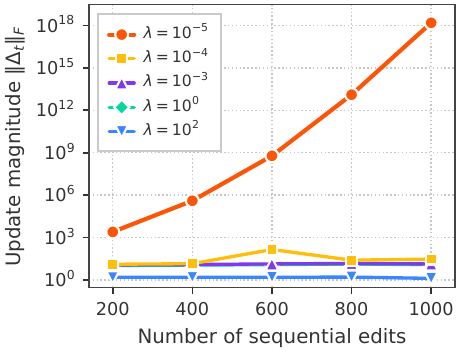}
    \caption{AlphaEdit update magnitude during sequential editing under different $\lambda$.}
    \label{fig:delta_norm}
\end{figure}

To complement the projection analysis in \S\ref{sec:rethinking}, we examine how AlphaEdit's weight updates evolve during sequential editing. Figure~\ref{fig:delta_norm} reports the total update magnitude per batch. For each edit batch, we aggregate the five edited layers by $\|\Delta_t\|_F=(\sum_{\ell}\|\Delta_t^{(\ell)}\|_F^2)^{1/2}$. It shows that AlphaEdit's update scale is sharply controlled by $\lambda$ during sequential editing. With extremely weak regularization ($\lambda=10^{-5}$), the update magnitude grows by many orders of magnitude as editing proceeds. At $\lambda=10^{-4}$, the trajectory no longer shows sustained exponential growth, but it exhibits large transient spikes, suggesting a near-unstable boundary. For $\lambda\ge 10^{-3}$, the update magnitude remains bounded throughout the edit stream. This weight-space view refines the main finding in \S\ref{sec:rethinking} that AlphaEdit's behavior is governed primarily by regularization strength, while the near-identity projector does not by itself prevent unstable updates.

\section{Full-Layer Visualization of the Null-Space Projector}
\label{app:p_layers}

\begin{figure}[t]
    \centering
    \includegraphics[width=\textwidth]{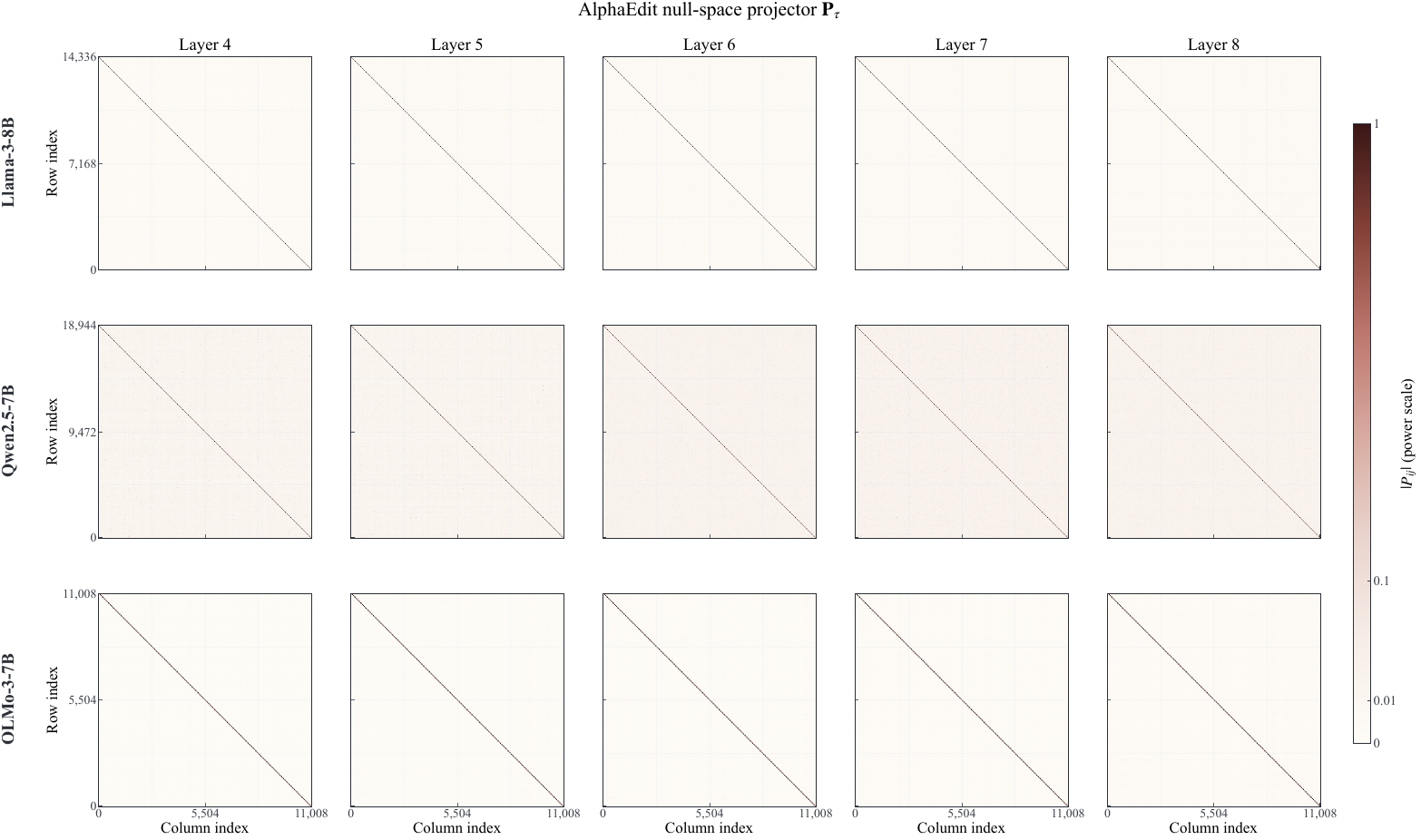}
    \caption{Full-layer visualization of AlphaEdit's null-space projector $\mathbf{P}_\tau$ across edited backbones.}
    \label{fig:app_p_layers_all}
\end{figure}

Figure~\ref{fig:app_p_layers_all} summarizes the full-layer projectors for the edited backbones. Across layers, $\mathbf{P}_\tau$ remains nearly diagonal and close to identity, indicating that AlphaEdit removes only a limited part of the activation space.

\section{Derivation of Eqn.~\eqref{eq:row_solution}}
\label{app:row_solution_derivation}

For the Fisher-weighted objective in Eqn.~\eqref{eq:goal}, define
\begin{equation}
\begin{split}
L(\tilde{\Delta}) =
&\|\tilde{\Delta}\mathbf{K}_1-\mathbf{E}\|^2
+\|\tilde{\Delta}\mathbf{K}_p\|^2 \\
&+\lambda\|\sqrt{\mathbf{F}}\odot\tilde{\Delta}\|^2 .
\end{split}
\end{equation}
Taking the derivative with respect to $\tilde{\Delta}$ gives
\begin{equation}
\begin{split}
\nabla_{\tilde{\Delta}} L
= {}&
2(\tilde{\Delta}\mathbf{K}_1-\mathbf{E})\mathbf{K}_1^\top
+2\tilde{\Delta}\mathbf{K}_p\mathbf{K}_p^\top \\
&+2\lambda(\mathbf{F}\odot\tilde{\Delta}) .
\end{split}
\end{equation}
Setting the derivative to zero and rearranging yields
\begin{equation}
\tilde{\Delta}
(\mathbf{K}_1\mathbf{K}_1^\top+\mathbf{K}_p\mathbf{K}_p^\top)
+\lambda(\mathbf{F}\odot\tilde{\Delta})
=
\mathbf{E}\mathbf{K}_1^\top .
\end{equation}
Let $\mathbf{G}=\mathbf{K}_1\mathbf{K}_1^\top+\mathbf{K}_p\mathbf{K}_p^\top$. Because the Fisher penalty is element-wise, this normal equation decomposes across rows. Let $\tilde{\delta}_i$, $e_i$, and $f_i$ denote the $i$-th rows of $\tilde{\Delta}$, $\mathbf{E}$, and $\mathbf{F}$, respectively, and define $\mathbf{D}_i=\mathrm{diag}(f_i)$. The $i$-th row therefore satisfies
\begin{equation}
\tilde{\delta}_i(\mathbf{G}+\lambda\mathbf{D}_i)
=
e_i\mathbf{K}_1^\top ,
\end{equation}
which gives the closed-form row-wise solution
\begin{equation}
\tilde{\delta}_i
=
e_i\mathbf{K}_1^\top(\mathbf{G}+\lambda\mathbf{D}_i)^{-1}.
\end{equation}
At the optimum, $\tilde{\delta}_i=\delta_i$, recovering Eqn.~\eqref{eq:row_solution}.

{For the efficient implementation, concatenate the current and historical keys as $\mathbf{U}=[\mathbf{K}_1\mid\mathbf{K}_p]\in\mathbb{R}^{m\times r}$, with $r=n_1+n_p$, so that $\mathbf{G}=\mathbf{U}\mathbf{U}^\top$. The Woodbury identity gives
\begin{equation}
\begin{split}
(\mathbf{G}+\lambda\mathbf{D}_i)^{-1}
= {}&(\lambda\mathbf{D}_i)^{-1}
-(\lambda\mathbf{D}_i)^{-1}\mathbf{U} \\
&\left(\mathbf{I}_r+\mathbf{U}^\top(\lambda\mathbf{D}_i)^{-1}\mathbf{U}\right)^{-1}
\mathbf{U}^\top(\lambda\mathbf{D}_i)^{-1}.
\end{split}
\label{eq:woodbury_inverse}
\end{equation}
Substituting this expression into Eqn.~\eqref{eq:row_solution} yields Eqn.~\eqref{eq:efficient_row_solution}, replacing the $m\times m$ inverse with diagonal scaling and an $r\times r$ linear solve.}

\section{Full Derivation of \S\ref{sec:theory}}
\label{app:proof}

In the following, we give the full derivation of the theoretical interpretation in \S\ref{sec:theory}. Specifically, we show that the element-wise Fisher regularizer in Eqn.~\eqref{eq:goal} corresponds to a computable diagonal approximation of the second-order Taylor expansion of the locality KL objective around $\Delta=0$, restricted to the edited parameter subspace.

\subsection{Setup and Notation}
\label{app:setup}
Let $\mathbf{W}$ denote the pre-edit value of the edited parameter block and $\Delta$ the update. Following Eqn.~\eqref{eq:loc}, the single-sample locality cost is
\begin{equation}
\ell_x(\Delta) = \mathrm{KL}\bigl(p_{\mathbf{W}}(\cdot\mid x)\,\|\,p_{\mathbf{W}+\Delta}(\cdot\mid x)\bigr),
\label{eq:ellx}
\end{equation}
and the reference-distribution objective is $\mathcal{L}_{\mathrm{loc}}(\Delta) = \mathbb{E}_{x\sim\mathcal{D}_{\mathrm{ref}}}[\ell_x(\Delta)]$.

According to Eqn.~\eqref{eq:ellx}, we have
\begin{equation}
\begin{aligned}
\ell_x(\Delta)
= &
\mathrm{KL}\bigl(p_{\mathbf{W}}(\cdot\mid x)\,\|\,p_{\mathbf{W}+\Delta}(\cdot\mid x)\bigr) \\
= &
\sum_y p_{\mathbf{W}}(y\mid x)
\log
\frac{p_{\mathbf{W}}(y\mid x)}{p_{\mathbf{W}+\Delta}(y\mid x)} \\
= &
\sum_y p_{\mathbf{W}}(y\mid x)\log p_{\mathbf{W}}(y\mid x) \\
 & - \sum_y p_{\mathbf{W}}(y\mid x)\log p_{\mathbf{W}+\Delta}(y\mid x).
\end{aligned}
\end{equation}
The first term depends only on the pre-edit distribution $p_{\mathbf{W}}(\cdot)$ and is therefore constant with respect to $\Delta$. Thus, we can write
\begin{equation}
\ell_{x}(\Delta)
\! = \!
\mathrm{const}(x)
 - \!
\sum_y \! p_{\mathbf{W}}(y \! \mid \! x)\log p_{\mathbf{W}+\Delta}\!(y \! \mid \! x),
\label{eq:ellx-split}
\end{equation}
where $\mathrm{const}(x)=\sum_y p_{\mathbf{W}}(y\mid x)\log p_{\mathbf{W}}(y\mid x)$. Therefore, only the cross-entropy term contributes derivatives with respect to $\Delta$.

\subsection{Regularity Conditions}
\label{app:regularity}
The derivation assumes the following standard regularity conditions on the model distribution:
\begin{itemize}
    \item \emph{Smoothness.} $p_{\mathbf{W}}(y\mid x)$ is twice continuously differentiable in the edited parameters $\mathbf{W}$.
    \item \emph{Parameter-independent support.} The support $\{y:p_{\mathbf{W}}(y\mid x)>0\}$ does not depend on $\mathbf{W}$. This holds for softmax outputs since every token has positive probability.
    \item \emph{Interchange of differentiation and summation/expectation.} Derivatives with respect to $w$ can be interchanged with the summation over $y$ and the expectation over $x\sim\mathcal{D}_{\mathrm{ref}}$.
    \item \emph{Finite Fisher moments.} $\mathbb{E}_{y\sim p_{\mathbf{W}}(\cdot\mid x)}
    \|\nabla_{\mathbf{W}}\log p_{\mathbf{W}}(y\mid x)\|^2<\infty$, and the corresponding Hessian expectation exists.
\end{itemize}
Under these conditions, the information equality between the KL Hessian and the Fisher information holds \citep{amari1998natural}.

\subsection{The Taylor Expansion of Single-Sample Locality Cost}
\label{app:first-order}

Next, we consider the Taylor expansion of Eqn.~\eqref{eq:ellx-split} around $\Delta=0$. Let $u=\mathrm{vec}(\Delta)$. We have,
\begin{equation}
\begin{aligned}
\ell_x(\Delta)
=&\
\ell_x(0)
+
\nabla_u \ell_x(0)^\top u  \\
&+
\tfrac{1}{2}u^\top
\nabla_u^2 \ell_x(0)
u + o(\|u\|^2).
\end{aligned}
\end{equation}

\paragraph{Zeroth order terms.} Setting $\Delta=0$ in Eqn.~\eqref{eq:ellx} gives $\ell_x(0) = \mathrm{KL}(p_{\mathbf{W}}\,\|\,p_\mathbf{W})=0$.

\paragraph{First order terms.} Differentiating Eqn.~\eqref{eq:ellx-split} with respect to $\Delta$ gives
\begin{equation}
\nabla_{\!\Delta}\ell_x(\Delta) \! = - \! \sum_y p_{\mathbf{W}}(y \! \mid \! x)\,\nabla_{\!\Delta}\log p_{\mathbf{W}+\Delta}(y \! \mid \! x),
\label{eq:supp_first_order}
\end{equation}

Let $\bar{\mathbf{W}}=\mathbf{W}+\Delta$. Since
$\mathrm{vec}(\bar{\mathbf{W}})=\mathrm{vec}(\mathbf{W})+\mathrm{vec}(\Delta)$,
the Jacobian $\partial \mathrm{vec}(\bar{\mathbf{W}})/\partial \mathrm{vec}(\Delta)$ is the identity. Therefore,
\begin{equation}
\begin{aligned}
& \left. \nabla_{\mathrm{vec}(\Delta)}\log p_{\mathbf{W}+\Delta}(y\mid x)
\right|_{\Delta=0} \\
= & \left. \frac{\partial \mathrm{vec}(\mathbf{W} \!\! + \!\! \Delta)}{\partial \mathrm{vec}(\Delta)} \nabla_{\mathrm{vec}(\mathbf{W} \! + \! \Delta)}
\log p_{\mathbf{W}+\Delta}(y \! \mid \! x)
\right|_{\Delta=0} \\
= & \nabla_{\mathrm{vec}(\mathbf{W})}\log p_{\mathbf{W}}(y\mid x).
\end{aligned}
\label{eq:supp_grad_delta}
\end{equation}
For $\Delta=0$, by Eqn.~\eqref{eq:supp_first_order} and Eqn.~\eqref{eq:supp_grad_delta}, we have
\begin{equation}
    \nabla_\Delta\ell_x(0) = -\sum_y p_{\mathbf{W}}(y\mid x)\,\nabla_{\mathbf{W}}\log p_{\mathbf{W}}(y\mid x).
    \label{eq:first_order_local}
\end{equation}

By swapping summation and differentiation, and using $\sum_y p_{\mathbf{W}}(y\mid x)=1$, we can derive
\begin{equation}
\begin{aligned}
     & \sum_y p_{\mathbf{W}}(y \mid x)\nabla_{\mathbf{W}} \log p_{\mathbf{W}}(y \mid x) \\
    = & \sum_y \nabla_{\mathbf{W}} p_{\mathbf{W}}(y \mid  x) \\
    = & \nabla_{\mathbf{W}} \sum_y p_{\mathbf{W}}(y \mid x) \\
    = & 0.
\end{aligned}
\end{equation}
Therefore, we have $\nabla_\Delta\ell_x(0)=0$.

\paragraph{Second order terms.}
Let $u:=\mathrm{vec}(\Delta)\in\mathbb{R}^{md}$ for $\Delta\in\mathbb{R}^{d\times m}$. Treat $\ell_x$ as a scalar function of $u$ and expand it around $u=0$. From the derivation above, both the zeroth-order and first-order terms vanish. Thus,
\begin{equation}
\ell_x(\Delta)
=
\tfrac{1}{2}\,u^\top \mathbf{H}_x\,u
+
o(\|u\|^2),
\label{eq:ellx-quadratic}
\end{equation}
where
\begin{equation}
\mathbf{H}_x
:=
\nabla_u^2\ell_x(\Delta)\big|_{\Delta=0}.
\end{equation}
To compute $\mathbf{H}_x$, we differentiate Eqn.~\eqref{eq:ellx-split} twice with respect to $u$. Since $p_{\mathbf{W}}(y\mid x)$ and $\mathrm{const}(x)$ do not depend on $\Delta$, we have
\begin{equation}
\nabla_u^2\ell_x(\Delta)
=
- \! \sum_y p_{\mathbf{W}}(y \! \mid \! x)\,
\nabla_u^2 \log p_{\mathbf{W}+\Delta}(y \! \mid \! x).
\end{equation}
Let $\bar{\mathbf{W}}=\mathbf{W}+\Delta$. As shown in Eqn.~\eqref{eq:supp_grad_delta}, $\mathrm{vec}(\bar{\mathbf{W}})$ depends on $u=\mathrm{vec}(\Delta)$ through an identity Jacobian. Therefore,
\begin{equation}
\left.
\nabla_u^2 \! \log p_{\mathbf{W}+\Delta}(y \! \mid \! x)
\right|_{\Delta=0}
\! = \!
\nabla_{\!\mathrm{vec}(\mathbf{W})}^2 \!
\log p_{\mathbf{W}}(y \! \mid \! x).
\end{equation}
Substituting this into the Hessian gives
\begin{equation}
\mathbf{H}_x
=
-\sum_y p_{\mathbf{W}}(y\mid x)\,
\nabla_{\mathrm{vec}(\mathbf{W})}^2
\log p_{\mathbf{W}}(y\mid x).
\label{eq:Hx}
\end{equation}
Define the single-sample Fisher information matrix restricted to the edited block as
\begin{equation}
\begin{aligned}
\mathbf{F}_x
:=
\sum_y p_{\mathbf{W}}(y\mid x)\,
\mathbf{s}_{\mathbf{W}}(y,x)
\mathbf{s}_{\mathbf{W}}(y,x)^\top ,
\end{aligned}
\label{eq:Fx}
\end{equation}
where
\begin{equation}
\mathbf{s}_{\mathbf{W}}(y,x)
:=
\nabla_{\mathrm{vec}(\mathbf{W})}
\log p_{\mathbf{W}}(y\mid x).
\end{equation}
Under the conditions in Appendix~\ref{app:regularity},
$\mathbf{H}_x=\mathbf{F}_x$. To see this, we differentiate
$\nabla_{\mathrm{vec}(\mathbf{W})}\log p_{\mathbf{W}}
=
p_{\mathbf{W}}^{-1}
\nabla_{\mathrm{vec}(\mathbf{W})}p_{\mathbf{W}}$
once more:
\begin{equation}
\begin{aligned}
& \nabla^2_{\mathrm{vec}(\mathbf{W})}\log p_{\mathbf{W}} \\
=
&\frac{
\nabla^2_{ \! \mathrm{vec}(\mathbf{W})}p_{\mathbf{W}}
}{
p_{\mathbf{W}}
} -
\frac{
\bigl(\nabla_{\! \mathrm{vec}(\mathbf{W})}p_{\mathbf{W}}\bigr)
\bigl(\nabla_{\!\mathrm{vec}(\mathbf{W})}p_{\mathbf{W}}\bigr)^\top
}{
p_{\mathbf{W}}^2
}.
\end{aligned}
\label{eq:logp-hessian}
\end{equation}
Multiplying Eqn.~\eqref{eq:logp-hessian} by
$p_{\mathbf{W}}(y\mid x)$ and summing over $y$ gives
\begin{equation}
\begin{aligned}
&-\sum_y p_{\mathbf{W}}(y\mid x)
\nabla^2_{\mathrm{vec}(\mathbf{W})}
\log p_{\mathbf{W}}(y\mid x) \\
=&
-\sum_y
\nabla^2_{\mathrm{vec}(\mathbf{W})}
p_{\mathbf{W}}(y\mid x) \\
&+
\!\!\sum_y \!
\frac{
\bigl(\nabla_{\! \mathrm{vec}(\mathbf{W})}
p_{\mathbf{W}}(y \! \mid \! x) \! \bigr)\!
\bigl(\nabla_{\! \mathrm{vec}(\mathbf{W})}
p_{\mathbf{W}}(y \! \mid \! x) \! \bigr)\!^\top
}{
p_{\mathbf{W}}(y \! \mid \! x)
}.
\end{aligned}
\end{equation}
The first term is zero because
\begin{equation}
\begin{aligned}
\sum_y
\nabla^2_{\mathrm{vec}(\mathbf{W})}
p_{\mathbf{W}}(y\mid x)
= &
\nabla^2_{\mathrm{vec}(\mathbf{W})}
\sum_y p_{\mathbf{W}}(y\mid x) \\
= &
\nabla^2_{\mathrm{vec}(\mathbf{W})}1 \\
= &
0.
\end{aligned}
\end{equation}
For the second term, using
$\nabla p_{\mathbf{W}}=p_{\mathbf{W}}\nabla\log p_{\mathbf{W}}$, we obtain
\begin{equation}
\begin{aligned}
&\sum_y
\frac{
\bigl(\nabla_{\mathrm{vec}(\mathbf{W})}
p_{\mathbf{W}}(y \! \mid \! x)\bigr)
\bigl(\nabla_{\mathrm{vec}(\mathbf{W})}
p_{\mathbf{W}}(y \! \mid \! x)\bigr)^\top
}{
p_{\mathbf{W}}(y \! \mid \! x)
} \\
=&
\sum_y p_{\mathbf{W}}(y \! \mid \! x)\,
\mathbf{s}_{\mathbf{W}}(y,x)
\mathbf{s}_{\mathbf{W}}(y,x)^\top
=
\mathbf{F}_x .
\end{aligned}
\end{equation}
Therefore,
\begin{equation}
\mathbf{F}_x
\! = \!
- \! \sum_y p_{\mathbf{W}}(y \! \mid \! x)
\nabla^2_{\! \mathrm{vec}(\mathbf{W})}
\log p_{\mathbf{W}}(y \! \mid \! x)
\! = \!
\mathbf{H}_x,
\label{eq:info-eq}
\end{equation}
which is the information equality.

\subsection{Averaging over the Reference Distribution}
\label{app:averaging}

Substituting $\mathbf{H}_x = \mathbf{F}_x$ into Eqn.~\eqref{eq:ellx-quadratic} and taking the expectation over $x\sim\mathcal{D}_{\mathrm{ref}}$,
\begin{equation}
\mathcal{L}_{\mathrm{loc}}(\Delta) \approx \tfrac{1}{2}\,\mathrm{vec}(\Delta)^\top \mathbf{F}^{(S)}\,\mathrm{vec}(\Delta),
\label{eq:loc-quadratic}
\end{equation}
where
\begin{equation}
\mathbf{F}^{(S)} := \mathbb{E}_{x\sim\mathcal{D}_{\mathrm{ref}}}[\mathbf{F}_x]
\label{eq:Fblock}
\end{equation}
is the Fisher information matrix \emph{restricted to the edited parameter subspace}. This restriction is not cosmetic: the edit only perturbs the weights of a single MLP down-projection block, so the Taylor expansion lives on that subspace and $\mathbf{F}^{(S)}$ is the corresponding Fisher block rather than the full-model Fisher.

\subsection{Diagonal Approximation and Recovery of Eqn.~\eqref{eq:goal}}
\label{app:diagonal}

For $\Delta\in\mathbb{R}^{d\times m}$, the full block $\mathbf{F}^{(S)}$ has size $(md)\times(md)$. At LLM scale ($md$ on the order of $10^7$), storing $\mathbf{F}^{(S)}$ or solving the associated quadratic program in closed form is infeasible. We therefore adopt the standard diagonal approximation
\begin{equation}
\mathbf{F}^{(S)} \approx \mathrm{Diag}(f), \qquad f\in\mathbb{R}^{md},
\end{equation}
and reshape $f$ into a matrix $\widetilde{\mathbf{F}} \in \mathbb{R}^{d\times m}$ of the same shape as $\Delta$. The quadratic form collapses to an element-wise penalty:
\begin{equation}
\mathrm{vec}(\Delta)^\top \mathrm{Diag}(f)\,\mathrm{vec}(\Delta) = \sum_{i,j} F_{ij}\,\Delta_{ij}^2.
\label{eq:diag-reshape}
\end{equation}
Combining Eqns.~\eqref{eq:loc-quadratic} and~\eqref{eq:diag-reshape},
\begin{equation}
\mathcal{L}_{\mathrm{loc}}(\Delta) \approx \tfrac{1}{2}\sum_{i,j} F_{ij}\,\Delta_{ij}^2,
\label{eq:loc-elementwise}
\end{equation}
which is exactly the regularizer appearing in Eqn.~\eqref{eq:goal}.

\section{Case Study}
\label{app:case_study}

We present representative paired outputs from Ripple. Each example reports two prompts from the same edited sample: a paraphrase of the edited fact and a neighborhood prompt about a nearby fact that should remain unchanged. Across these paired examples, FisherEdit consistently produces the edited answer while preserving the unrelated local fact.

\begin{tcolorbox}[
    enhanced,
    breakable,
    colback=white,
    colframe=black!15,
    boxrule=0.4pt,
    arc=1mm,
    left=4pt,
    right=4pt,
    top=4pt,
    bottom=4pt,
    title=\textbf{Case 1},
    colbacktitle=black!3,
    coltitle=black,
    fonttitle=\small,
    before skip=4pt,
    after skip=6pt
]
\small
\noindent\textbf{Edit / Neighborhood.} John Masly: occupation \emph{researcher} $\rightarrow$ \emph{unemployed}; gender should remain \emph{male}.

\noindent\textbf{Outputs.} FisherEdit: \emph{unemployed} / \emph{male}. AlphaEdit: \emph{unknown} / \emph{unknown}. AnyEdit: ``John Masly is a historian.'' / ``John P Masly is male.'' MEMIT: unrelated edit continuation / ``male and female farmers, also genderfarmersqueepea''.
\end{tcolorbox}

\begin{tcolorbox}[
    enhanced,
    breakable,
    colback=white,
    colframe=black!15,
    boxrule=0.4pt,
    arc=1mm,
    left=4pt,
    right=4pt,
    top=4pt,
    bottom=4pt,
    title=\textbf{Case 2},
    colbacktitle=black!3,
    coltitle=black,
    fonttitle=\small,
    before skip=4pt,
    after skip=6pt
]
\small
\noindent\textbf{Edit / Neighborhood.} Sharifah Mazlina binti Syed Abdul Kadir: occupation \emph{explorer} $\rightarrow$ \emph{music professor}; citizenship should remain \emph{Malaysia}.

\noindent\textbf{Outputs.} FisherEdit: \emph{music professor} / \emph{Malaysia}. AlphaEdit: ``music teacher and calligrapher'' / \emph{Iran}. AnyEdit: ``Minister of Women's Empowerment and Family Development in'' / ``citizen of Malaysia.'' MEMIT: corrupted text on both prompts.
\end{tcolorbox}

\begin{tcolorbox}[
    enhanced,
    breakable,
    colback=white,
    colframe=black!15,
    boxrule=0.4pt,
    arc=1mm,
    left=4pt,
    right=4pt,
    top=4pt,
    bottom=4pt,
    title=\textbf{Case 3},
    colbacktitle=black!3,
    coltitle=black,
    fonttitle=\small,
    before skip=4pt,
    after skip=6pt
]
\small
\noindent\textbf{Edit / Neighborhood.} Anthony Deakin: occupation \emph{researcher} $\rightarrow$ \emph{art dealer}; alma mater should remain \emph{University of Liverpool}.

\noindent\textbf{Outputs.} FisherEdit: \emph{art dealer} / \emph{University of Liverpool}. AlphaEdit: \emph{art dealer} / ``known only to a few''. AnyEdit: ``Royal Navy officer and politician.'' / \emph{University of Sheffield}. MEMIT: degenerate continuations on both prompts.
\end{tcolorbox}

\clearpage
\section{Future Work}
\label{app:future_work}

\paragraph{Scaling to longer edit streams.}
An important direction is to extend FisherEdit to substantially longer editing sequences by improving the scalability of its solver and historical-key memory.
In Eqn.~\eqref{eq:efficient_row_solution}, $r=n_1+n_p$ counts both the current and accumulated historical edit keys, and the row-wise solve costs $O(dmr^2+dr^3)$ per edited weight matrix.
Since Algorithm~\ref{alg:fisheredit} appends new keys after each batch, $n_p$ grows throughout the stream even when the batch size is fixed.
The computational advantage of the Woodbury formulation relies on $r\ll m$, motivating further improvements to the row-wise solver as historical constraints accumulate.
Future work could explore incremental updates to the row-wise factorizations and warm-started iterative solvers that reuse computations across successive edit batches, with the aim of reducing editing latency while preserving earlier edits.

\paragraph{Knowledge editing in reasoning models.}
FisherEdit's parameter-importance regularization provides a natural basis for updating factual knowledge while protecting the reasoning processes that use it.
The method requires locate-then-edit MLP locations and a Fisher diagonal estimated on a calibration set, without architectural modifications.
The calibration set serves as an adaptable control knob: estimating the Fisher diagonal directly on complex reasoning traces, such as multi-step mathematical or logical trajectories, would penalize changes to parameters with high sensitivity on these traces.
This offers a way to direct preservation toward reasoning-related behavior while allowing factual updates through less sensitive parameters.
Future work can investigate this separation between factual recall and reasoning, and develop sequential-editing evaluations that jointly assess successful knowledge replacement, the use of updated facts in multi-step solutions, and preservation of reasoning ability.

\paragraph{Complementarity with retrieval and deployment scenarios.}
Parameter editing and retrieval-based approaches are complementary.
When retrieved context conflicts with memorized knowledge, models may over-rely on stale parametric memory~\citep{longpre2021entity}, and their use of external evidence depends on its coherence and persuasiveness~\citep{xie2024adaptive}.
Editing updates the fact at its source, so the correction can remain available when retrieval misses, ranks poorly, is only partially relevant, or is disabled.
This motivates combining FisherEdit with retrieval-augmented generation or agentic search to improve consistency between parametric knowledge and external evidence.
On-device, offline, and privacy-sensitive deployments also provide useful settings for exploration, particularly when maintaining a retrieval corpus and pipeline is impractical.
Once applied, FisherEdit introduces no additional inference-time computation; latency- and cost-sensitive serving can therefore benefit from factual corrections that do not require a retrieval step on every query.

\end{document}